\documentclass[10pt,twocolumn,letterpaper]{article}

\usepackage{cvpr}              

\usepackage{graphicx}
\usepackage{amsmath}
\usepackage{amssymb}
\usepackage{booktabs}
\usepackage{multirow}
\usepackage{algorithm}
\usepackage[noend]{algpseudocode}
\usepackage{sidecap}
\usepackage[dvipsnames]{xcolor}

\newcommand{\Name}{PMGAN}

\usepackage[pagebackref,breaklinks,colorlinks]{hyperref}

\usepackage[capitalize]{cleveref}
\crefname{section}{Sec.}{Secs.}
\Crefname{section}{Section}{Sections}
\Crefname{table}{Table}{Tables}
\crefname{table}{Tab.}{Tabs.}

\def\cvprPaperID{6519} 
\def\confName{CVPR}
\def\confYear{2022}

\begin{document}

\title{Polymorphic-GAN: Generating Aligned Samples across Multiple Domains with Learned Morph Maps}

\title{\vspace{-6mm}Polymorphic-GAN: Generating Aligned Samples across Multiple Domains with Learned Morph Maps \vspace{-5mm}}

\author{Seung Wook Kim $^{1,2,3}$  \quad\quad Karsten Kreis$^{1}$ \quad\quad Daiqing Li$^{1}$ 
\and
  \quad Antonio Torralba$^{4}$    \quad Sanja Fidler$^{1,2,3}$ \\
  \small{\textsuperscript{1}NVIDIA \quad \textsuperscript{2}University of Toronto \quad \textsuperscript{3}Vector Institute \quad \textsuperscript{4}MIT \vspace{1pt}}\\
  \; \quad \texttt{\scriptsize \{seungwookk,kkreis,daiqingl,sfidler\}@nvidia.com,} \texttt{\scriptsize   torralba@mit.edu}\\
  \url{https://nv-tlabs.github.io/PMGAN/}\\
}

\twocolumn[{%
\renewcommand\twocolumn[1][]{#1}%
\maketitle
\begin{center}
\vspace{-7.5mm}
    \centering
    \captionsetup{type=figure}
    \includegraphics[width=1\textwidth]{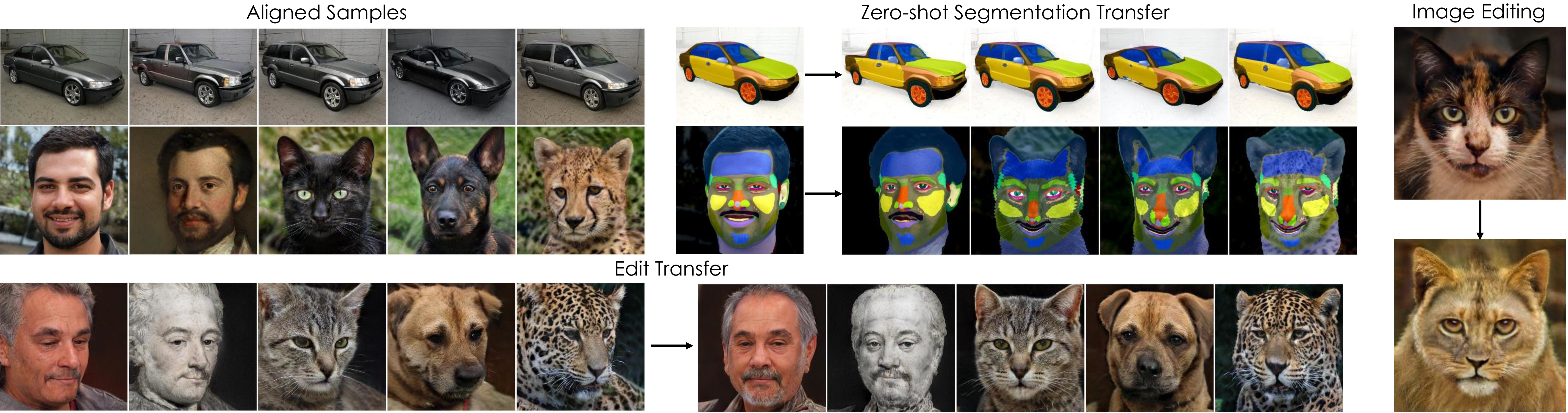}
    \vspace{-7mm}
    \captionof{figure}{We introduce \emph{Polymorphic-GAN} ({\Name}) for generating aligned samples across multiple domains. PMGAN enables a diverse set of applications, including zero-shot segmentation and cross-domain image editing, by learning geometric differences between domains.}
\end{center}%
}]
\begin{abstract}
    Modern image generative models show remarkable sample quality when trained on a single domain or class of objects. 
    In this work, we introduce a 
    generative adversarial network that can simultaneously generate aligned image samples from multiple related domains.
    We leverage the fact that a variety of object classes share common attributes, with certain geometric differences. 
    We propose Polymorphic-GAN which learns shared features across all domains and a per-domain morph layer to morph shared features according to each domain.
    In contrast to previous works, our framework allows simultaneous modelling of images with highly varying geometries, such as images of human faces, painted and artistic faces, as well as multiple different animal faces.
    We demonstrate that our model produces aligned samples for all domains and show how it can be used for 
    applications such as segmentation transfer and cross-domain image editing, as well as training in low-data regimes.   
    Additionally, we apply our Polymorphic-GAN on image-to-image translation tasks and show that we can greatly surpass previous approaches in cases where the geometric differences between domains are large.

\end{abstract}


\vspace{-5.0mm}
\section{Introduction}
\label{sec:intro}
\vspace{-0.5mm}

Generative adversarial networks (GANs) have achieved remarkable image synthesis quality~\cite{karras2019style,karras2020analyzing,esser2021taming,brock2018large}. Moreover, GANs like StyleGAN~\cite{karras2019style,karras2020analyzing} have been shown to form a semantic understanding of the modeled images in their features~\cite{goetschalckx2019ganalyze,bau2019semantic,bau2019gandissect,jahanian2020Osteerability,voynov2020unsupervised,harkonen2020ganspace,shen2021closed,tritrong2021repurposing,li2021semantic,zhang21,xu2021linear}, which has been leveraged in diverse applications, including image editing~\cite{collins2020editing,zhu2020domain,hou2020guidedstyle,kim2021drivegan,kim2021stylemapgan,tov2021designing,ling2021editgan}, inverse rendering~\cite{zhang2020image}, style transfer~\cite{kazemi2019style,abdal2019image2stylegan}, image-to-image translation~\cite{isola2017image,choi2018stargan,choi2020starganv2,richardson2021encoding}, and semi-supervised learning~\cite{li2021semantic,zhang21,xu2021linear}.

GANs are usually trained on images from individual domains, such as human faces~\cite{karras2017progressive,karras2019style}. However, there are many related domains which share similar semantics and characteristics, such as animal faces or face paintings.
In our work, we aim to train a generative model with a shared backbone to produce \emph{aligned} samples from multiple related domains. By aligned, we mean images that share common attributes and conditions across domains, such as pose and lighting. This has an obvious computational advantage by sharing weights across domains, but more importantly, it affords a variety of applications such as transferring segmentation labels from one domain to another, in which such information may not be available. Furthermore, by editing one domain, we get edits in other domains for free. 

The main obstacle to building a GAN that simultaneously synthesizes outputs from different domains is that even though the \textit{semantics} are often shared, the \textit{geometry} can vary significantly (consider, for example, the face of a human, a dog, and a cat). This prevents a natural sharing of generator features among such semantically aligned, but geometrically varying domains. Common approaches such as fine-tuning a pre-trained GAN~\cite{mo2020freeze,Karras2020ada,wu2021stylespace} unfortunately lose the ability to sample from the parent domain, or, more generally, multiple domains at the same time. Learning shared representations between multiple domains has been studied in the transfer and multi-task learning literature~\cite{dai2016instance,zhao2018modulation,liu2019loss,ma2018modeling}, but there has been little progress in generative models~\cite{liu2017unsupervised,huang2018multimodal,liu2019few,choi2020starganv2,baek2021rethinking}.

To overcome these challenges, we propose \textit{Polymorphic-GAN} ({\Name}). It leverages a shared generator network together with novel \textit{morph maps} that geometrically deform and adapt the synthesis network's feature maps. In particular, {\Name} builds on the StyleGAN2~\cite{karras2020analyzing} architecture and augments the model with a \emph{MorphNet} that predicts domain-specific morph maps which warp the main generator's features according to the different domains' geometries. An additional shallow convolutional neural network is then sufficient to render these morphed features into correctly stylized and geometrically aligned outputs that are also semantically consistent across multiple domains. 

By sharing as many generator layers as possible, the impressive semantic properties of StyleGAN's latent space are shared across all modeled domains, while the geometric differences are still correctly reflected due to the additional morph operations. Because of that, our {\Name} enables many relevant applications in a unique and novel way. We extensively analyze {\Name} and validate the method on the following tasks: \textit{(i)} We perform  previously impossible expressive image editing across different domains. 
\textit{(ii)} We use {\Name} for image-to-image translation across domains, and outperform previous methods 
in cases where the geometric gap between domains is large. 
\textit{(iii)} We leverage {\Name}'s learnt morph maps 
for zero-shot semantic segmentation transfer across domains. 
\textit{(iv)} Finally, sharing the generator's features across domains is advantageous when involving domains with little training data. In these cases the main generator can be learnt primarily from a domain with much data, which benefits all other domains.
In summary, our {\Name} is the first generative model that naturally and easily allows users to synthesize aligned samples from multiple semantically-related domains at the same time, enabling novel and promising applications.

\vspace{-3mm}
\section{Related Work}
\label{sec:related_work}

\textbf{StyleGAN.}
StyleGAN~\cite{karras2019style,karras2020analyzing} is the state-of-the-art GAN model with remarkable sample quality, which has enable many relevant applications. 
StyleGAN inversion methods~\cite{richardson2021encoding,abdal2019image2stylegan,tov2021designing,zhu2020domain} discover the latent vector corresponding to an input image. 
Once an image is embedded,
GAN-based editing methods~\cite{shen2020interfacegan,wu2021stylespace,shen2021closed,ling2021editgan} find semantically meaningful directions in latent space to achieve desired editing effects.
DatasetGAN~\cite{zhang21} and SementicGAN~\cite{li2021semantic} use StyleGAN's feature maps for producing segmentation masks.

\textbf{StyleGAN Adaptation.}
Various methods adapt a pre-trained StyleGAN for a target domain.
Fine-tuning approaches~\cite{mo2020freeze, wu2021stylespace, Karras2020ada} take a pre-trained model as a starting point for optimization and learn to generate samples from the target domain.
Few-shot adaptation approaches~\cite{ojha2021few,patashnik2021styleclip,gal2021stylegan} make use of a small amount of data from the target domain or CLIP~\cite{radford2021learning} to adapt latent codes or model weights 
towards the target domain.
{\Name}, in contrast, learns a single generator that jointly models multiple domains.

\textbf{Cross-Domain Generation.}
Several works~\cite{liu2017unsupervised,huang2018multimodal,saito2020coco,liu2019few,choi2018stargan,choi2020starganv2,lee2018diverse,mao2019mode,zhu2017unpaired,park2020contrastive} learn image-to-image translation across a pair or several domains. UNIT~\cite{liu2017unsupervised} and MUNIT~\cite{huang2018multimodal} learn shared representations between two domains for image translation. StarGAN~\cite{choi2018stargan,choi2020starganv2} learns a single network that takes in an input image and a style code to translate the input to multiple domains. SemanticGAN~\cite{li2021semantic} jointly produces images and segmentation masks.
On top of these, our model allows unique applications by exploiting geometry.

\textbf{Leveraging Geometry.}
Keypoint representations have been used to find landmarks in an unsupervised way~\cite{mokady2021jokr,thewlis2017unsupervised,zhang2018unsupervised,suwajanakorn2018discovery,wu2019transgaga}. 
TransGaGa~\cite{wu2019transgaga} uses a conditional VAE~\cite{kingma2013auto} to  learn a heat map of facial landmarks to aid the image translation task.
Jaderberg \etal~\cite{jaderberg2015spatial} proposes a differentiable module to spatially modify feature maps.
DeepWarp~\cite{ganin2016deepwarp} learns to warp images for gaze manipulation.
Caricature generation~\cite{cao2018cari,shi2019warpgan,gong2020autotoon} has benefitted from warping the input photo for exaggerated facial features.
However, they require supervision through paired caricature and photo data or facial feature detectors.
In contrast, our generative model jointly models multiple domains by learning the geometric differences in a completely unsupervised manner.

\section{Polymorphic-GAN}
\label{sec:method}
\vspace{-1mm}
In Sec.~\ref{sec:motivation} we describe the motivation for our proposed approach,  the model architecture in Sec.~\ref{sec:pretrained_stylegan} to ~\ref{sec:morphing} and the training procedure in Sec.~\ref{sec:training}.


\begin{SCfigure*}
  \centering
   \caption{{\Name} generates aligned samples across multiple domains. All domains share the core StyleGAN generator. MorphNet produces domain-specific morph maps and warps the generator features to be geometrically suitable for the target domains. The learned morph maps can be exploited for interesting applications as we demonstrate in this work. We train the model with domain-specific discriminators.}
{\vspace{-1mm}\includegraphics[width=0.71\textwidth]{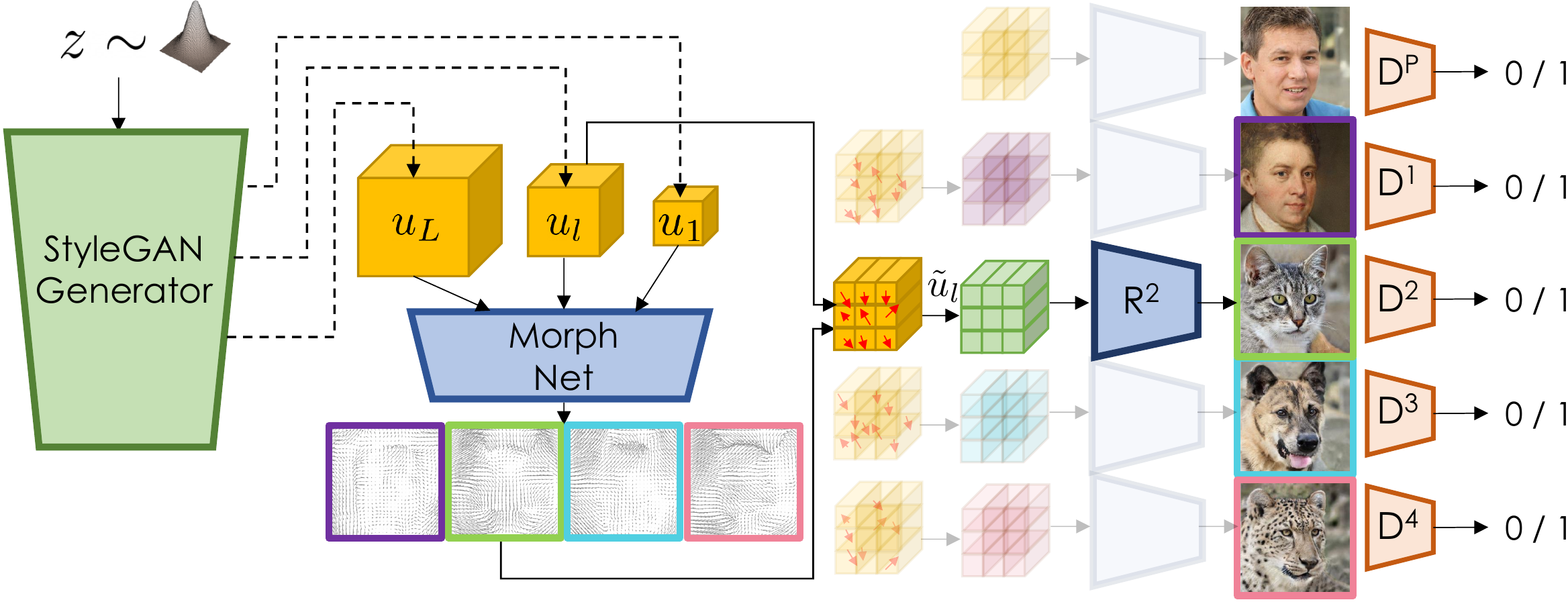}}
   \label{fig:model_diagram}
  \end{SCfigure*}
  
\subsection{Motivation}
\label{sec:motivation}
\vspace{-1mm}
We aim to learn a GAN-based generator
that can simultaneously synthesize \textit{aligned} images from multiple different domains. 
We denote images as \textit{aligned} if they share common attributes and conditions across domains, such as pose and lighting condition. 
To train such a model, it is critical that the features of the generator can be translated into each domain, and by sharing more layers in the generator, we can naturally enforce such alignment better.
The intermediate features in modern GAN~\cite{karras2020analyzing,karras2019style,brock2018large} generators are generally shaped as 3-dimensional tensors with spatial dimensions.
They go through rendering layers such as shallow convolution layers to produce an output image.

Consider two image domains such as human faces and portrait paintings. 
They share many attributes, including geometry, and we can easily model both domains with shared generator layers and small domain-specific convolution layers that render facial features according to their domain. 
However, suppose the domains have a more significant gap in geometry, such as human and cat faces. 
In this case, the rendering layers have to render geometrically different facial landmarks at different positions. 
The generator needs to learn the geometric differences by interpreting features in different ways for each domain, which discourages feature sharing. Therefore, we either need to reduce the number of shared layers in the generator or increase the number of domain-specific layers, either of which is not desirable as \textit{alignment} between domains will be less enforced.

To overcome this problem, 
our {\Name} utilizes a domain-specific morph net that learns in a fully unsupervised manner the geometric differences between domains and morphs the shared features for each domain. 
It allows the sharing of entire generator layers across all domains while still having only shallow domain-specific rendering layers.
We note that, in concurrent work, Wu \etal~\cite{wu2021stylealign} analyzes how finetuning of a pre-trained StyleGAN2 model from a parent to a child domain affects the model's network weights.
The paper shows that the weights of the convolution layers in the main generator change the most, in addition to the mapping layers that produce the style vectors, which also change noticeably, especially when the geometric gap between the domains is large. These strong network parameter changes that are required to adapt the model indicate that naive feature sharing between domains, particularly geometrically different domains, is highly non-trivial. This is in line with our hypothesis. Our {\Name} more efficiently allows sharing of features across domains by explicitly modelling geometry in the generator's feature space.

\subsection{Pre-trained StyleGAN}
\label{sec:pretrained_stylegan}
\vspace{-1mm}
{\Name} is based on StyleGAN2~\cite{karras2020analyzing} and we extend it to multiple domains (Figure~\ref{fig:model_diagram}).
We denote the set of datasets for domains to be trained as $\mathcal{D} = \{\pi^P, \pi^1, ..., \pi^N\}$ where $\pi^P$ is a special dataset from the parent domain for which we assume that there exists a StyleGAN2 model pre-trained on $\pi^P$. 
{\Name} is composed of the pre-trained StyleGAN2's generator $G$, domain-specific morph layers $M^{1,...,N}$ and rendering layers $R^{1,...,N}$.
We first sample a noise vector $z \sim p(z)$ from the standard Normal prior distribution and feed it through $G$, 
which produces the output image $I^P$ and also the intermediate features $u_1, ..., u_L$ for $L$ features in $G$.
Specifically, we store the generator features for each spatial resolution from $2^2\times2^2$ to $2^{L+1}\times2^{L+1}$ before the final features are transformed via a $1\times1$ convolution layer (\ie tRGB) that produces the output RGB values. We assume square images with $H=W=2^{L+1}$.

\subsection{MorphNet}
\label{sec:morphnet}
\vspace{-1mm}
Features $u_1, ..., u_L$ contain valuable information, including semantic content as well as fine-grained edge information.
We use these features to produce domain-specific morph maps that can modify the geometry embedded in the features to be suitable for each target domain.
The MorphNet component of {\Name} first reduces each feature map's channel dimension to be smaller through a $1\times1$ convolution layer and then upsamples all features to match the largest spatial resolution $H\times W$. 
The upsampled features are concatenated channel-wise and go through two $3\times3$ convolution layers.
We add a fixed 2-dimensional sinusoidal positional encoding~\cite{vaswani2017attention} to the merged features to inject grid position information which can be useful for learning geometric biases in a dataset.  
Finally, this tensor is processed by domain-specific convolutional layers $M^d$ for each domain $d$. 
$M^d$ produces a $H\times W\times2$ morph map $\mathcal{M}^d_\Delta$, normalized between $[-1/\eta, 1/\eta]$ through Tanh activation function where $\eta$ is a hyperparameter that controls the maximum displacement we allow the morphing operation to produce.
$\mathcal{M}^d_\Delta$ represents the relative horizontal and vertical direction that each pixel would get its value from (a pixel here is $(p,q)$ position in a 3-dim spatial tensor).

\begin{algorithm}
\caption{Inference step for {\Name}}\label{alg:inf}
\begin{algorithmic}
\Function{forward}{$z$}
\State $u_1, ..., u_L, \mathcal{I}^{P}= G(z)$
\State $u = \texttt{MergeFeatures}(u_1, ..., u_L)$
\For{$d \in (1, ..., N)$}
    \State $\mathcal{M}_\Delta^d = M^d(u) $ \Comment{Get morph map for domain $d$} 
    \State $\{\tilde{u}_1, ..., \tilde{u}_L\}_d = \texttt{Morph}(u_l,\mathcal{M}_\Delta^d)$ \textbf{for all} $l$
    \State $\mathcal{I}^{d} = R^d(\tilde{u}_1, ..., \tilde{u}_L)$ 
\EndFor
\State \textbf{return} $\mathcal{I}^1, ..., \mathcal{I}^N, \mathcal{I}^P$
\EndFunction
\end{algorithmic}
\end{algorithm}

\begin{figure}
  \centering
   \includegraphics[width=1.0\linewidth]{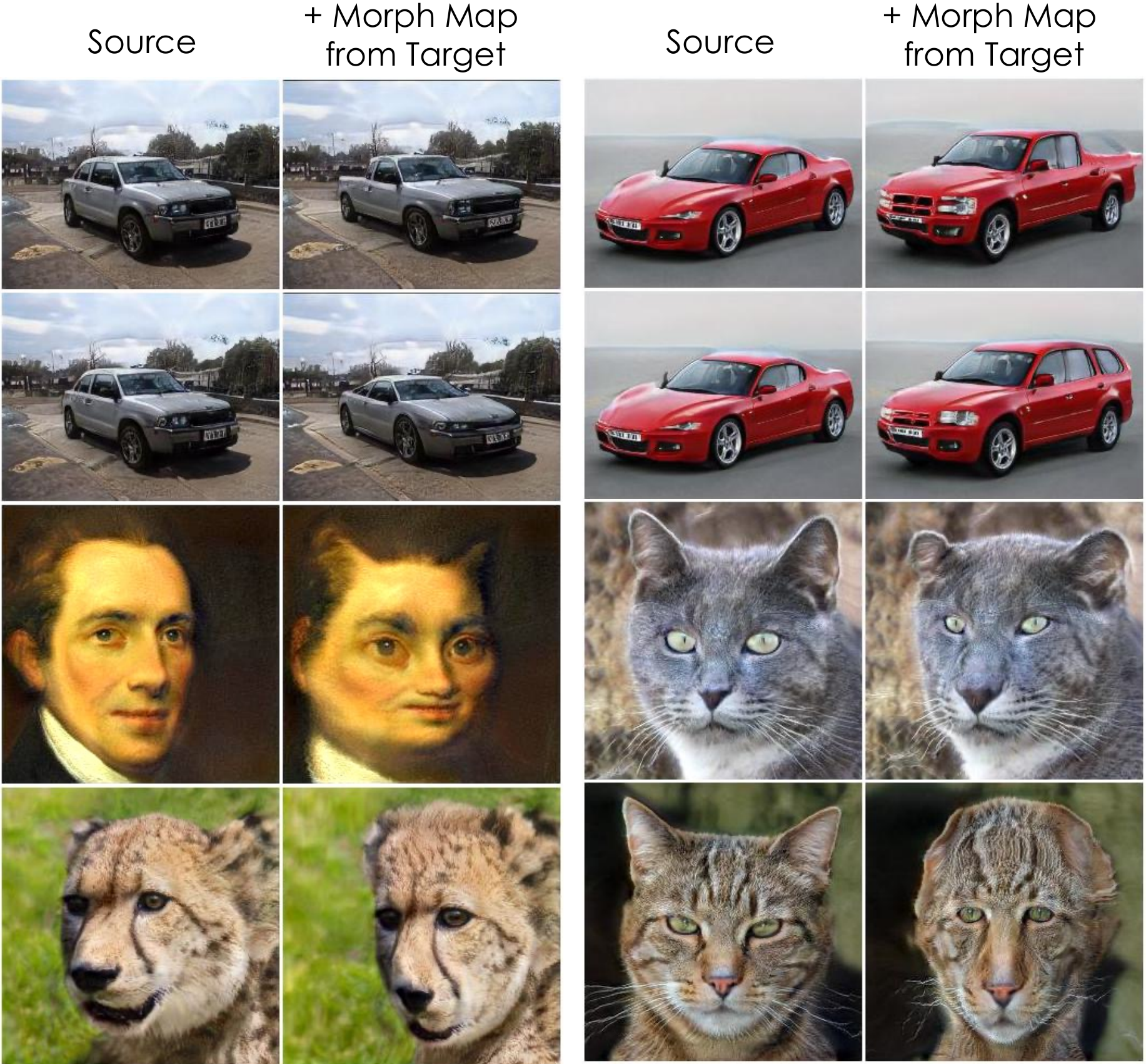}
\vspace{-6mm}
   \caption{Effect of using target domain's morph map while keeping everything else fixed in the source domain.}
   \label{fig:morph_only}
   \vspace{-2mm}
\end{figure}
 
\subsection{Feature Morphing}
\label{sec:morphing}
\vspace{-1mm}
We follow Spatial Transformer Networks (SPN)~\cite{jaderberg2015spatial} to differentiably morph features with $\mathcal{M}^d_\Delta$. 
We initialize a 2D sampling grid from an identity transformation matrix, normalized between $[-1, 1]$.
The sampling grid has the same shape as $\mathcal{M}^d_\Delta$, and each pixel $(p,q)$ in the sampling grid contains the absolute position $(x,y)$ of the source pixel that will be morphed into $(p,q)$. For example, if pixel $(p,q)$ has value $(-1, -1)$, the vector at the top left corner of the source feature map will be morphed into $(p,q)$.
The morph map $\mathcal{M}_{\Delta}^d$ is added to the grid, and we denote the resulting grid as $\Gamma \in \mathbb{R}^{H\times W\times 2}$. 
Unlike SPN that produces an affine transformation matrix with six parameters for sampling grid, we learn pixel-wise morphing maps, which gives us precise control for fine-detailed morphing. 
For each layer $l$ of generator features, we perform the following \texttt{Morph} operation that bilinearly interpolates features: 
\begin{equation}
    \tilde{u}_l^{pq} = \sum_{n}^{H_l}\sum_{m}^{W_l} u_l^{nm}\text{max}(0, 1-|x^{pq} -m |)\text{max}(0,1-|y^{pq}-n|)
\end{equation}
where $\tilde{u}_l^{pq}$ is the morphed feature vector at pixel $(p,q)$ for layer $l$, $u_l^{nm}$ is the source feature vector prior to \texttt{Morph} at pixel $(n,m)$, and $(x^{pq}, y^{pq})$ is the sample point in $\Gamma$ for pixel $(p,q)$, assuming unnormalized grid coordinates for ease of presentation. 
Note that $\Gamma$ is also bilinearly interpolated to match the spatial dimension of each layer $(H_l, W_l)$.

The morphed features $\{\tilde{u}_1, ..., \tilde{u}_L\}_d$ are now geometrically transformed to be suitable for domain $d$.
Each of these features is then processed via further convolution layers $R^d$ to produce RGB images. 
They are finally summed together using skip connections as in StyleGAN2~\cite{karras2020analyzing}. 
Importantly, the $R^d$ layers can correct small unnatural distortions caused by the feature morphing process, in contrast to previous works that directly warp output images~\cite{shi2019warpgan}.

The inference step of {\Name} is summarized in Algo.~\ref{alg:inf}. 

\subsection{Training}
\label{sec:training}
\vspace{-1mm}
We use separate discriminators with the same architecture for each domain, and train {\Name} with non-saturating logistic loss~\cite{goodfellow2014generative}, R1 regularization~\cite{mescheder2018training} and path-length regularization~\cite{karras2020analyzing}. 
We use equal loss weightings for all domains, 
except when we do low-data regime training (Sec.~\ref{sec:low_data}), in which case we weigh losses by $|\pi^d|/\text{max}_l|\pi^l|$ where $|\pi^d|$ is the number of training examples in domain $d$.
The intuition is that we want the generator features to be mostly learned from data-rich domains while 
domains with significantly less data leverage the rich representation with domain-specific layers.
The StyleGAN2 generator is initialized from pre-trained weights on a parent domain. 
We found that initializing all discriminators from the same pre-trained model helped stabilize training. 
We freeze the first three layers of discriminators and the shared generator, and do not update these weights~\cite{mo2020freeze,Karras2020ada}.
We also share the weights of $k$ rendering layers of $R$ across domains which promotes rendering of similar style such as colors. 
The more rendering layers we share, the more similar in style domains become, but that comes with the tradeoff of not being able to learn domain-specific styles.
We found setting $k=1$ or sharing the rendering layer at $4\times 4$ spatial resolution was adequate in producing similarly styled outputs. 
We set the morph hyperparameter $\eta=3$ for all experiments such that each pixel can move at most $1/6$ of the image size in the $x$ and $y$ direction. 
$\eta$ can be adjusted depending on the geometric gap between domains. 


\section{Experiments}
\label{sec:experiments}

\begin{figure*}[!thb]
  \centering
\includegraphics[width=0.9\textwidth]{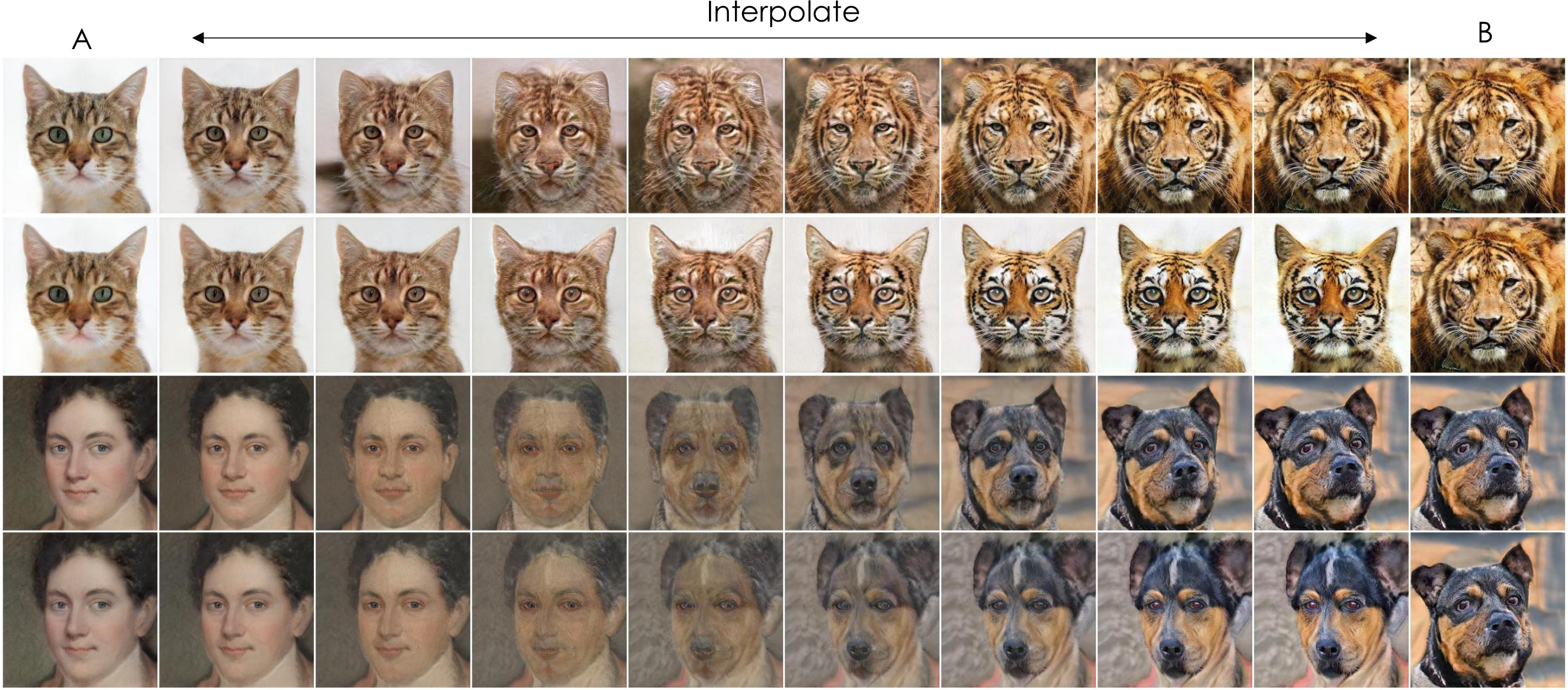}
   \vspace{-2mm}
   \caption{ 
   \textbf{Odd Rows:} Linearly interpolating both domain-specific layers of two domains and their latent vectors A\&B.
   \textbf{Even Rows:} Linearly interpolating domain-specific layers and latent vectors while keeping the morph map of A the same.}
   \vspace{-2mm}
\label{fig:morphing_face}
\end{figure*}

\noindent We construct two multi-domain datasets for evaluation: 

\textbf{Cars} dataset consists of five classes of cars from the LSUN-Car dataset~\cite{yu15lsun}. We use an object classifier by Ridnik \etal~\cite{ridnik2021imagenet21k} that can output fine-grained object classes to divide the dataset into the following domains: Sedan (149K), SUV (52K), Sports car (58K), Van (25K), and Truck (22K), with the number of images in parentheses.
The parent StyleGAN2 model is pre-trained on Sedan.

\textbf{Faces} dataset consists of Flickr-Faces-HQ~\cite{karras2019style} (70K), MetFaces~\cite{Karras2020ada} (1.3K), as well as Cat (5.6K), Dog (5.2K) and Wild life (5.2K) from the AFHQ dataset~\cite{choi2020starganv2}.
The parent model is pre-trained on the Flickr-Faces-HQ dataset. 

AFHQ datasets have official testing splits, and we use 5\% of the other domains as testing sets. 

We carry out all experiments at 256$\times$256 resolution. 
The datasets contain domains with varying geometric differences. 
Our goal is to learn both large and subtle geometric and texture differences between them.
We provide more details on each section in the supplementary materials.



\subsection{Ablation Studies}
\label{sec:sample_quality}
\vspace{-1mm}
\begin{table}
\centering
\large
\resizebox{\linewidth }{!}{ 
\begin{tabular}{l|c|ccccc}

    \toprule
    Criterion & Method        & Sedan & Truck & SUV & Sports Car & Van \\
    \midrule
   \multirow{4}{*}{FID ($\downarrow$)}
    &  *DC-StyleGAN2 & 18.0 &  120.1 & 189.1 & 80.1 & 111.3 \\
    & DC-StyleGAN2 & 10.3 & 93.4 & 82.6 & 93.3 & 96.5 \\
    & Ours w.o Morph & 5.7 & 35.7 & 18.5 & 16.0 & 34.9\\
    & Ours  & \textbf{5.4} & \textbf{23.3} & \textbf{11.3} & \textbf{9.1} & \textbf{19.3} \\
    \midrule
    \multirow{4}{*}{Acc. ($\uparrow$)} 
   & *DC-StyleGAN2 & 85.9\% & 2.4\% & 6.3\% & 21.5\% & 16.2\%  \\
   & DC-StyleGAN2 & 65.8\%& 3.8\%& 27.3\%& 14.0\%& 22.6\%\\
   & Ours w.o Morph & 84.7\% & 45.2\% & 56.3\% & 75.3\% & 42.3\% \\
   & Ours          & \textbf{88.2\%} & \textbf{69.2\%} & \textbf{74.9\%} & \textbf{88.8\%} & \textbf{73.9\%} \\

    \bottomrule
\end{tabular}
}
\vspace{-3mm}
\caption{
FID and classification accuracy for Cars dataset.
}
\label{tab:lsun_baseline}
\vspace{-2mm}
\end{table}

\begin{table}
\centering
\large
\resizebox{\linewidth }{!}{ 
\begin{tabular}{l|c|ccccc}
    \toprule
    Criterion & Method        & FFHQ & Metfaces & Cat & Dog & Wild Life \\
    \midrule
     \multirow{4}{*}{FID ($\downarrow$)}
    & *DC-StyleGAN2 & \textbf{6.6} & 46.3 & 127.4 & 66.4 & 102.0 \\
    & DC-StyleGAN2  &14.3 & 47.9 & 16.0 & 60.7 & 18.1 \\
    & Ours w.o Morph & 8.1 & 37.5 & 21.3 & 63.8 & 27.8\\
    & Ours          & 7.4 & \textbf{34.7} & \textbf{9.4} & \textbf{34.5} & \textbf{12.0} \\
     \midrule
    \multirow{4}{*}{Acc. ($\uparrow$)} 
    & *DC-StyleGAN2 & 99.7\% & 87.8\% & 11.8\% & 77.0\% & 41.8\% \\
    & DC-StyleGAN2 & 92.5\% & 81.7\% & 96.2\% & 83.8\% & 87.3\% \\
    & Ours w.o Morph & \textbf{99.9\%} & \textbf{100.0\%} & 96.6\% & 94.2\% & 98.8\% \\
    & Ours          & \textbf{99.9\%} & \textbf{100.0\%} & \textbf{99.5\%} & \textbf{98.6\%} & \textbf{99.7\%} \\
    \bottomrule
\end{tabular}}
\vspace{-3mm}
\caption{
FID and classification accuracy for Faces dataset.
}
\label{tab:faces_baseline}
\vspace{-2mm}
\end{table}

We first verify the efficacy of {\Name} on producing aligned samples across domains with a single model.
Our first baseline \textit{Domain-Conditional StyleGAN2 (DC-StyleGAN2)} is a modified StyleGAN2 model that takes a one-hot encoded domain vector as an input. 
The one-hot vector is embedded through a linear layer, concatenated with the output of the mapping network, merged with a linear layer and fed into the generator. 
\textit{*DC-StyleGAN2} has the same architecture as \textit{DC-StyleGAN2}, but it starts from a pretrained model and only adds the extra layers for class-conditioning.
The next one is \textit{Ours without Morph}, the same as our full PMGAN, except for the MorphNet component that morphs the generator features.  

We measure sample quality with Fr\'echet Inception Distance (FID)~\cite{heusel2017gans} and accuracy using pretrained domain classifiers. 
The classifiers measure if models produce corresponding samples for each domain. 
They are implemented as ResNet-18~\cite{he2016deep} for the 5-way classification task.

\begin{figure}
  \centering
   \includegraphics[width=1.0\linewidth]{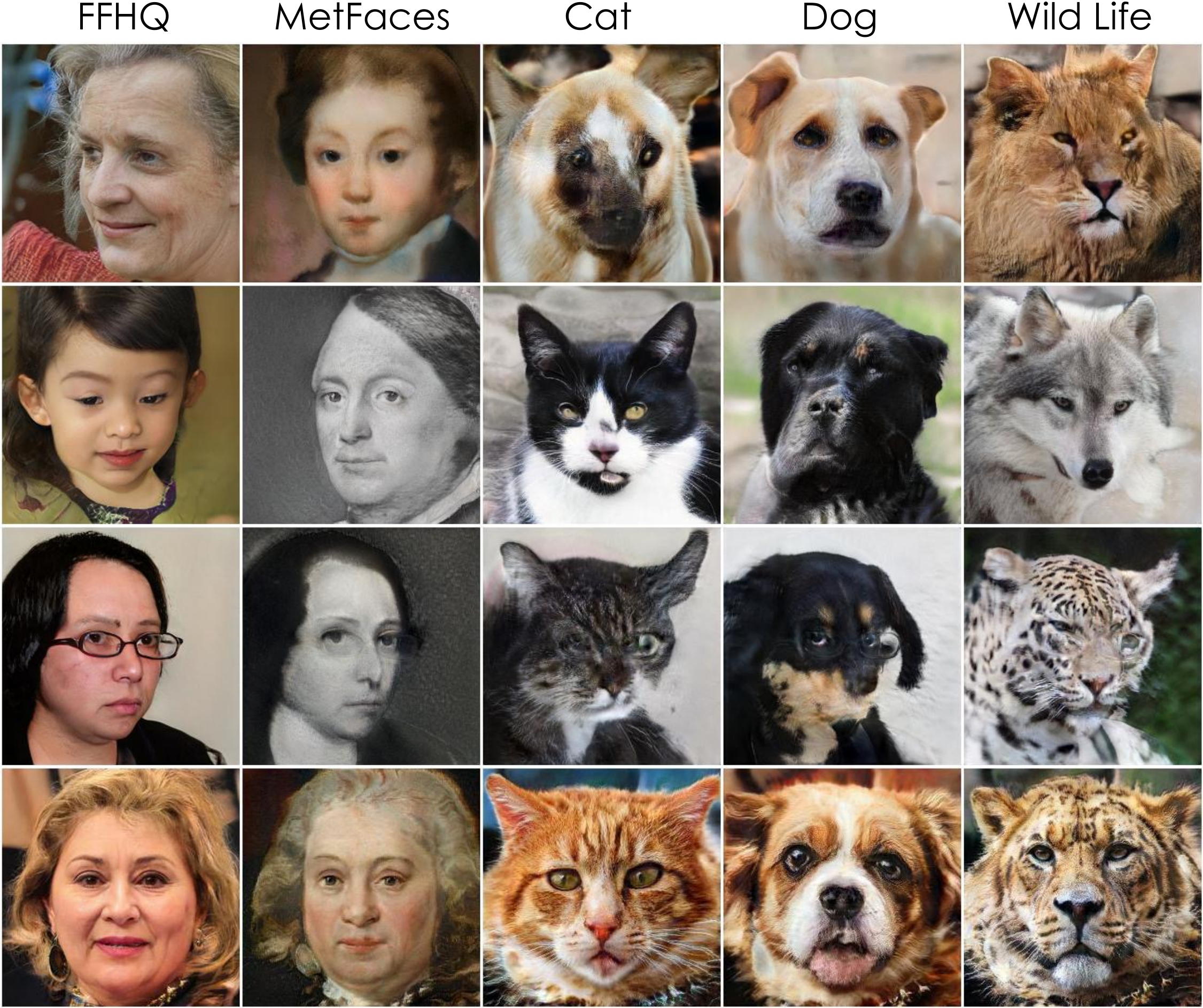}
\vspace{-7mm}
   \caption{Each row is a sample from one latent vector. \emph{Top row}: *DC-StyleGAN2, \emph{Second row}: DC-StyleGAN2, \emph{Third row}: Ours without Morphing, \emph{Last row}: Our {\Name}.}
   \label{fig:baseline}
\end{figure}

Tab.~\ref{tab:lsun_baseline}, \ref{tab:faces_baseline} show ablation results. 
\textit{DC-StyleGAN2} produces reasonable samples but they are not aligned well across domains as the domain conditioning in the generator modifies generator features to be specialized for each domain, as can be seen in Fig.~\ref{fig:baseline}. 
We found that \textit{*DC-StyleGAN2}, which finetunes a pre-trained model, has difficulties learning class-conditioning information, as can be seen in its low classification accuracy. We suspect that it is not trivial to adapt generator features to be suitable across domains without any domain-specific layers.
\textit{Ours without Morph} produces aligned poses as it tries to use the same generator features, but has trouble sharing features across domains because of their geometric differences. 
It cannot effectively use the shared features, as features corresponding to certain facial landmarks, such as eyes, nose and mouth, often vary in spatial position across domains.

In contrast, our full PMGAN leverages domain-specific layers but still benefits from sharing the entire stack of features due to the geometric morphing. It achieves the best overall sample quality and accuracy on both datasets. 

\vspace{-1mm}
\subsection{Qualitative Analysis}
\label{sec:qualitative}
\vspace{-1mm}

We start by analyzing what {\Name} has learned in the morph maps.
Similar to model interpolation~\cite{wang2019deep,wu2021stylealign}, we can interpolate the domain-specific layers of {\Name} to continuously interpolate two different domains. 
Additionally, we leverage the morph maps to investigate if the models correctly learned the geometric differences. 
We sample two latent vectors A and B, and linearly interpolate domain layer weights as well as the latent vectors.
As can be seen in Fig.~\ref{fig:morphing_face}, {\Name} is capable of performing cross-domain interpolation, and by fixing A's morph map during interpolation, it maintains the geometric characteristic of A while adapting to B's texture.
This shows how geometry is disentangled from rendering and {\Name} can be used for interesting image editing applications 
such as transforming a cat to look like a tiger. 
In Fig.~\ref{fig:morph_only}, we show the effect of using the target domain $t$'s morph map for a source domain $s$. Specifically, we swap the morph map  $\mathcal{M}^s_\Delta$ with $\mathcal{M}^t_\Delta$ and render for domain $s$. 
For Cars, whose domains have similar texture, we can see how cars from source domains can be smoothly transformed towards the target domain. 
For Faces, we see interesting rendering such as a cat-shaped human face.
These results demonstrate how {\Name} successfully learned the distinct geometries of each domain.

\begin{figure}
  \centering
   \includegraphics[width=1.0\linewidth]{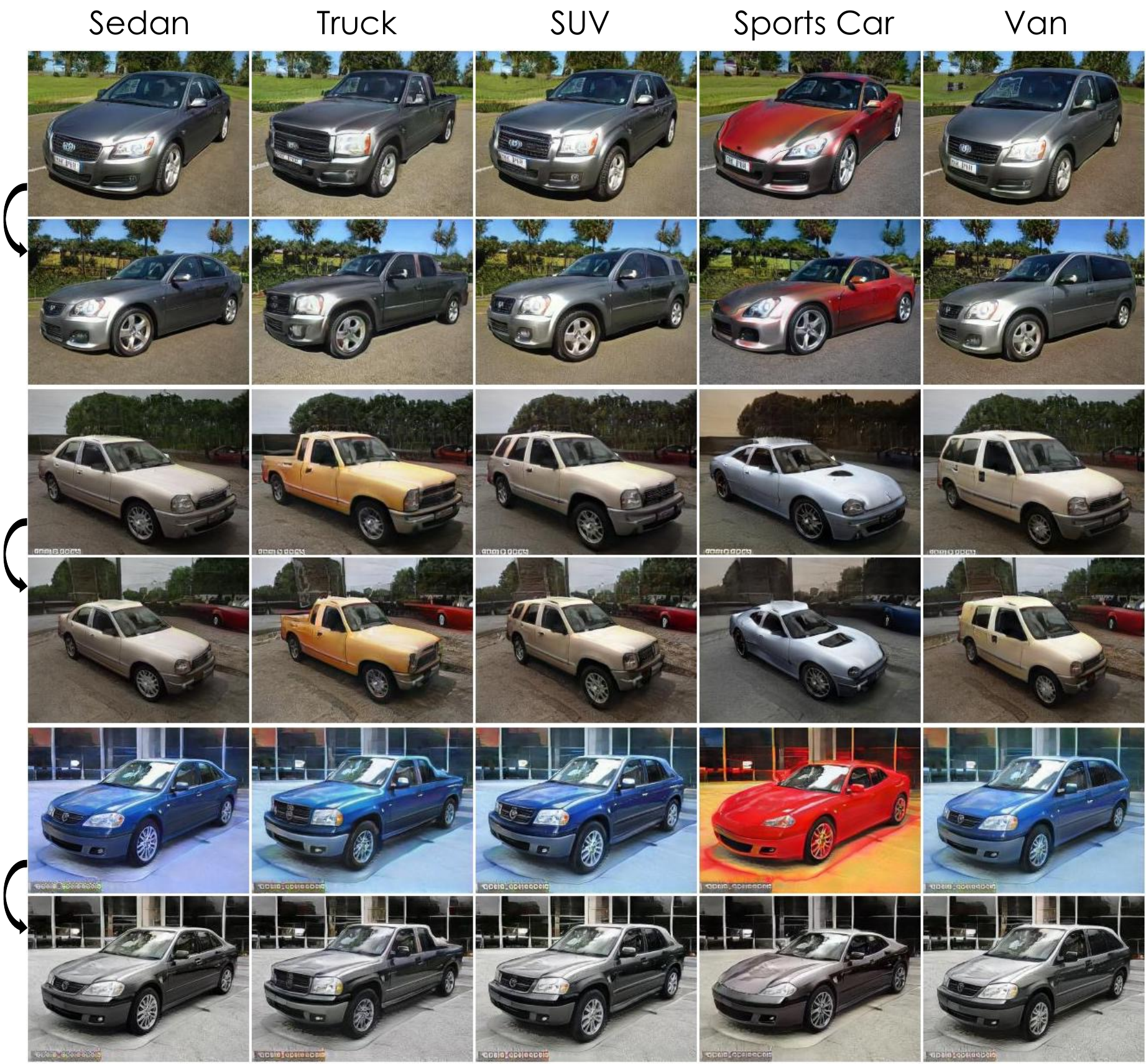}
\vspace{-6.5mm}
   \caption{Edit transfer. Edit directions discovered through {\Name}'s core generator can be transferred across all domains. Top: rotation, Middle: zoom, Bottom: color.}  
   \label{fig:lsun_edit}
   \vspace{-3mm}
\end{figure}

\begin{table}
\centering
\large
\resizebox{\linewidth }{!}{ 
\begin{tabular}{l|c|cccc|c}
    \toprule
    Criterion & Method  & Truck & SUV & Sports Car & Van & Mean \\
    \midrule
   \multirow{2}{*}{mIoU ($\uparrow$)}
    & Baseline & 0.45 & 0.57 & 0.52 & 0.44 & 0.49 \\
    & Ours  & \textbf{0.67} & \textbf{0.74} & \textbf{0.63} &  \textbf{0.64} & \textbf{0.67} \\

    \bottomrule
\end{tabular}}
\vspace{-3mm}
\caption{
Mean IoU for zero-shot segmentation. Our transferred segmentation masks show high IoU with pseudo-labelled masks.
}
\vspace{-2mm}
\label{tab:seg_trans}
\end{table}

\begin{figure}
  \centering
   \includegraphics[width=1.0\linewidth]{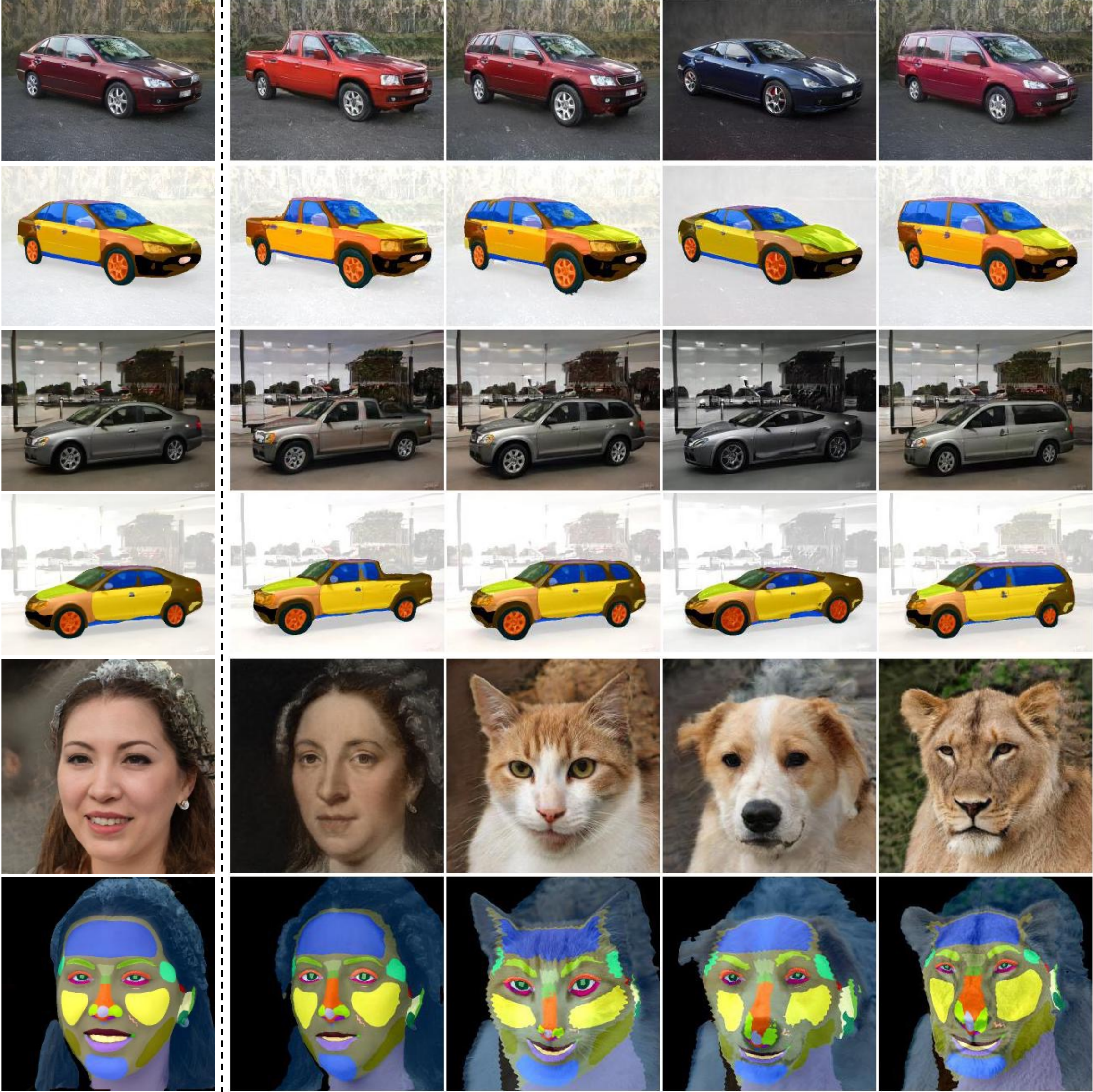}
\vspace{-6.5mm}
   \caption{Zero-shot segmentation transfer. The masks in the left-most column are transferred to other domains using $\mathcal{M}_\Delta$.}
   \label{fig:seg_transfer}
   \vspace{-3mm}
\end{figure}

\begin{SCfigure*}
  \centering
\includegraphics[width=0.82\textwidth]{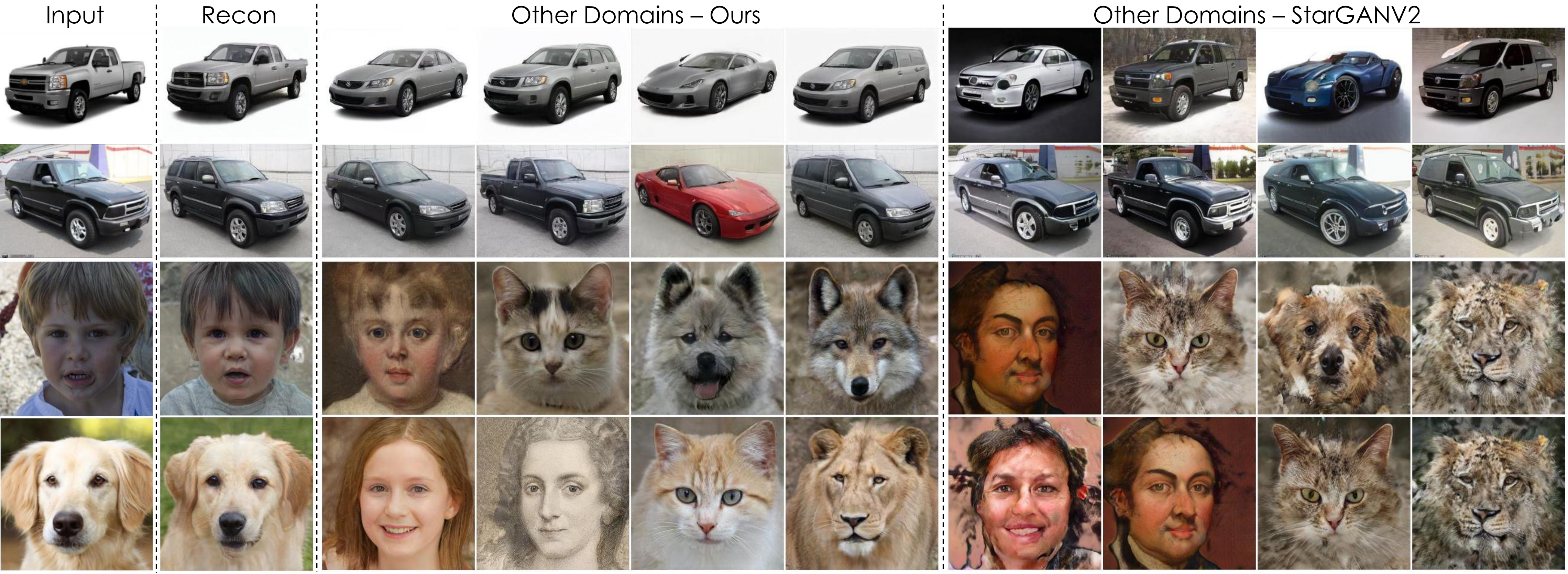}
   \vspace{-3mm}
   \caption{ 
   We compare image to image translation results from {\Name} and StarGANv2. {\Name} uses GAN inversion techniques to find a latent vector that can reconstruct the input image and renders the other domains.}
   \vspace{-1mm}
\label{fig:stargan_comparison}
\end{SCfigure*}

\textbf{Edit Transfer}
There has been tremendous interest~\cite{shen2020interfacegan,wu2021stylespace,shen2021closed,ling2021editgan} in disentangling StyleGAN's latent space to find useful \textit{edit vectors} that can modify the output image in a semantically meaningful way by pushing the latent vector of StyleGAN into certain directions.
{\Name}'s aligned cross-domain samples through the shared generator allow us to discover edit vectors
that transfer across domains.
We use SeFa~\cite{shen2021closed} for its simplicity to find edit vectors in {\Name}.
We find meaningful vectors such as rotation, zoom, lighting and elevation.
Fig.~\ref{fig:lsun_edit} shows some examples of how edit vectors can be transferred across all domains.



\vspace{-1mm}
\subsection{Zero-shot Segmentation Transfer}
\label{sec:seg_transfer}
\vspace{-1mm}
Assuming there exists a method that can output a segmentation map for images from the parent domain, it is possible to zero-shot transfer the segmentation mask to all other domains using {\Name}'s learned morph map.
We directly use the \texttt{Morph} operation on the segmentation map with $\mathcal{M}_\Delta$ after bilinearly interpolating the morph map to match the size of the mask.
As the morph map $\mathcal{M}_\Delta$ captures the geometric differences between domains, we can successfully use $\mathcal{M}_\Delta$ to transfer the parent's segmentation masks across domains as shown in Fig.~\ref{fig:seg_transfer}.
To measure the quality of the segmentation transfer, we use a pre-trained segmentation network to pseudo-label detailed car parts following Zhang \etal~\cite{zhang21}.
We compare the agreement between the pseudo-label and transferred segmentations from the sedan class. 
In Tab.\ref{tab:seg_trans}, the baseline measures mean IoU using the segmentation from the sedan class for all classes without morphing, which serves as a good baseline as {\Name} produces aligned samples whose poses are mostly identical. 
Our zero-shot segmentation shows much higher agreement with the pseudo-label, indicating our model correctly learned the correspondence between different car parts.

\begin{figure*}[!htb]
  \centering
 \includegraphics[width=0.99\textwidth]{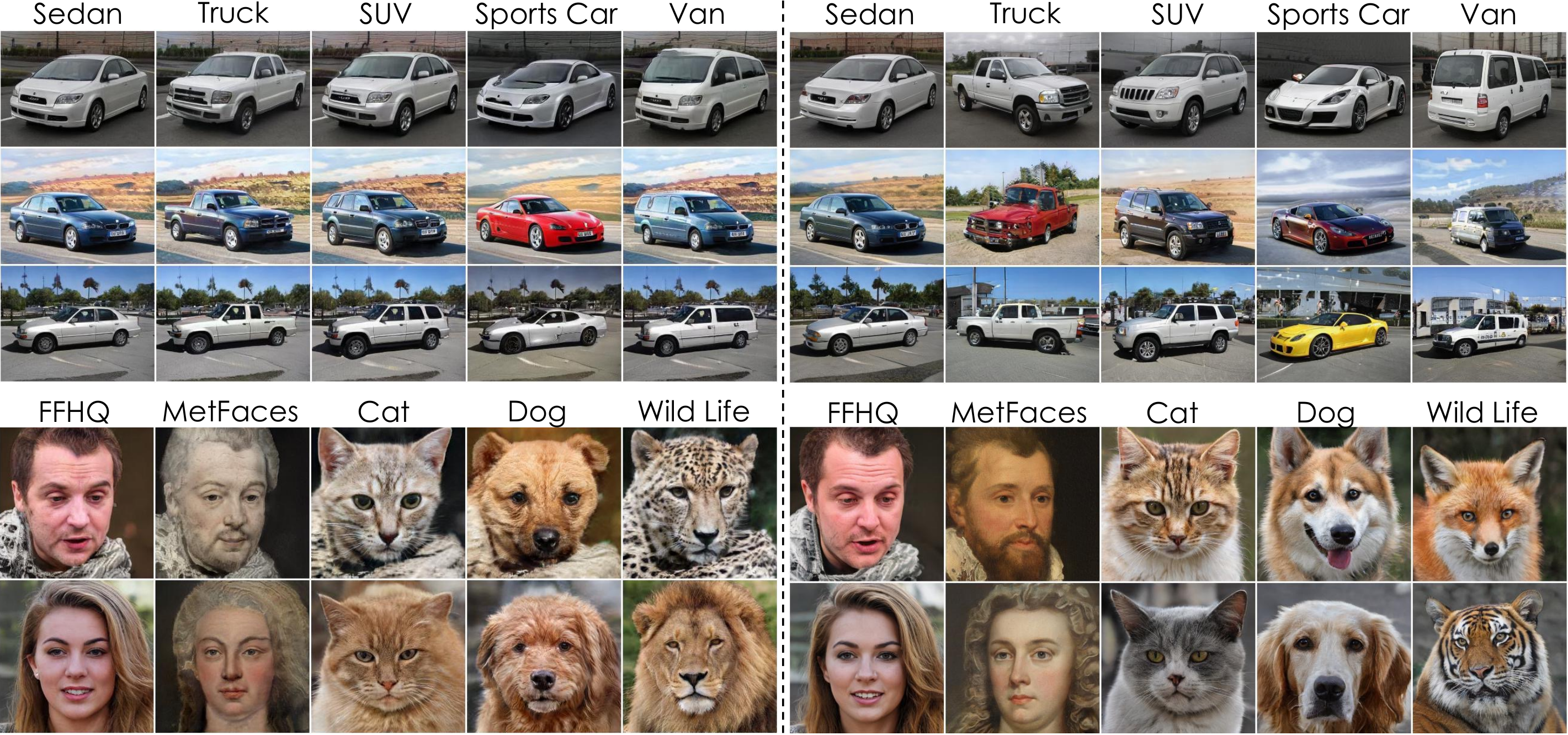}
 \vspace{-2mm}
   \caption{\textbf{Left:} Samples from {\Name}, \textbf{Right:} Samples from finetuned models from the same parent model. Our model produces consistently aligned samples across domains. Finetuning~\cite{mo2020freeze}  specializes models for each domain, especially for less pre-aligned datasets.}
\label{fig:finetune_compare}
\vspace{-3mm}
\end{figure*}

\vspace{-1mm}
\subsection{Image-to-Image Translation}
\label{sec:i2i}
\vspace{-1mm}
There is a large body of work that does \textit{GAN inversion}~\cite{richardson2021encoding,abdal2019image2stylegan,tov2021designing,zhu2020domain} with StyleGAN.
{\Name} can easily use any GAN inversion method as the model is based on StyleGAN. 
Once an image is inverted in the latent space, {\Name} can naturally be used for image-to-image translation (I2I) tasks by synthesizing every other domain with the same latent code.
On the Cars dataset, we use latent optimization~\cite{karras2019style} to encode input images, and outperform the state-of-the-art multi-domain image translation model StarGANv2~\cite{choi2020starganv2} on both FID and accuracy. 
StarGANv2 has a strong shape bias from the input image and has trouble translating to another car domain, as indicated by its low accuracy in Tab.~\ref{tab:lsun_i2i}. 
For the Faces dataset, StarGANv2 does well if trained only on animal faces because geometric differences between animal classes are small. However, when trained on all five domains of Faces, training collapses and fails to translate between human and animal faces (Tab.~\ref{tab:faces_i2i} and Fig.~\ref{fig:stargan_comparison}).
To compare with other image translation approaches, we also evaluate on a single domain translation task in Tab.~\ref{tab:single_i2i}. 
{\Name} shows competitive performance on the Cat-to-Dog task, despite being a generative model trained on all five domains together, as opposed to methods that only translate between two domains (except StarGANv2, which models three animal domains together).

\begin{table}
\centering
\large
\resizebox{\linewidth }{!}{ 
\begin{tabular}{l|c|ccccc|c}
    \toprule
    Criterion & Method        & Sedan & Truck & SUV & Sports Car & Van & Mean \\
    \midrule
   \multirow{2}{*}{FID ($\downarrow$)}
    & StarGANv2~\cite{choi2020starganv2} & 28.1 & 35.0 & 41.0 & \textbf{20.7} & 42.2 & 33.4 \\
    & Ours  & \textbf{26.7} & \textbf{25.3} & \textbf{26.6} & 25.6 & \textbf{35.9} & \textbf{28.0} \\
    \midrule
    \multirow{2}{*}{Acc. ($\uparrow$)} 
   & StarGANv2~\cite{choi2020starganv2} & 48.5\% & 62.2\% & 58.1\% & 84.4\% & 63.5\% & 63.3\% \\
   & Ours          & \textbf{94.1\%} & \textbf{90.0\%} & \textbf{80.0\%} & \textbf{91.6\%} & \textbf{76.2\%} & \textbf{86.4}\% \\
    \bottomrule
\end{tabular}
}
\vspace{-3mm}
\caption{
I2I performance on Cars. Each column evaluates quality of samples translated from other domains to the column's domain.}

\label{tab:lsun_i2i}
\vspace{-2mm}
\end{table}

\begin{table}
\centering
\large
\resizebox{\linewidth }{!}{ 
\begin{tabular}{l|ccccc}
    \toprule
   Evaluation Dataset        & MUNIT~\cite{huang2018multimodal} & DRIT~\cite{lee2018diverse} & MSGAN~\cite{mao2019mode} & StarGANv2~\cite{choi2020starganv2} & Ours \\
    \midrule
    Animals Only & 41.5 & 95.6 & 61.4 & \textbf{16.2} & 33.1 \\
    All Domains  & - & - & - & 133.7 & \textbf{41.1} \\
    \bottomrule
\end{tabular}}
\vspace{-3mm}
\caption{
I2I performance on Faces (FID). Ours is trained on all domains from Faces for both rows. Other models are trained only on animals for the first row, and on all domains for the second row.
}
\label{tab:faces_i2i}
\vspace{-2mm}
\end{table}

\begin{table}
\centering
\large
\resizebox{\linewidth }{!}{ 
\begin{tabular}{l|cccccc}
    \toprule
    Task        & MUNIT~\cite{huang2018multimodal} & CycleGAN~\cite{zhu2017unpaired} & StarGANv2\cite{choi2020starganv2} & CUT~\cite{park2020contrastive}  &Ours   \\
    \midrule
    Cat$\rightarrow$ Dog & 91.4 & 76.3 & \textbf{53.4} &56.4 & 55.9\\
    \bottomrule
\end{tabular}}
\vspace{-3mm}
\caption{
I2I performance on Cat-to-Dog (FID). 
Ours shows competitive performance despite being a generative model that jointly models all domains.
}
\label{tab:single_i2i}
\vspace{-2mm}
\end{table}


\vspace{-1mm}
\subsection{Low Data Regime}
\label{sec:low_data}
\vspace{-1mm}

\begin{table}
\centering
\large
\resizebox{\linewidth }{!}{ 
\begin{tabular}{l|c|ccc}
    \toprule
    Dataset        & Method & 5\% & 20\% & 100\% \\
    \midrule
    \multirow{2}{*}{Metfaces} 
    & StyleGAN2~\cite{karras2020analyzing} - single domain &  68.7 & 83.0 & 72.4 \\
    & Ours - five domains &  59.7&  40.7 & 34.7 \\
    \midrule
    \multirow{2}{*}{AFHQ-Cat} 
    & StyleGAN2~\cite{karras2020analyzing} - single domain & 27.3 & 19.6 & 6.8\\
    & Ours - five domains & 23.3 & 13.8 & 9.4  \\    
    \bottomrule
\end{tabular}}
\vspace{-3mm}
\caption{
Low data regime (FID). For different amount of data used, we compare ours with StyleGAN2 trained on a single domain.
}
\label{tab:low_data}
\vspace{-2mm}
\end{table}

{\Name} shares features for multiple domains, which can be beneficial for domains with small amounts of data, as they can leverage the rich representations learned from other domains.
We evaluate {\Name} on the Faces dataset by varying the amount of data for the MetFaces and Cat domains while other domains use the full training data (Tab.~\ref{tab:low_data}). 
Compared to StyleGAN2, we achieve better FIDs when the amount of training data is small.
Note that StyleGAN2 training with 5\% data mode-collapsed. However, FID was not robust enough to reflect this, as the number of data in MetFaces (1.3K) is too small.
{\Name} can be combined with techniques that explicitly tackle low-data GAN training~\cite{Karras2020ada,zhao2020diffaugment,Sauer2021NEURIPS}, which we leave for future work.


\vspace{-1mm}
\subsection{Comparison to Plain Fine-Tuning}
\vspace{-1mm}


There have been recent works~\cite{gal2021stylegan,wu2021stylealign,patashnik2021styleclip,ojha2021few} on fine-tuning pre-trained StyleGANs for new target domains. While these methods can achieve high image quality, fine-tuning encourages the child models to be specialized to the new domains. 
As a further comparison, we fine-tune the same parent model used by our PMGAN for each domain~\cite{mo2020freeze}.
For Faces, fine-tuning preserves some attributes such as pose and colors (with the same latents for original and fine-tuned models), but Fig.~\ref{fig:finetune_compare} shows {\Name} achieves better alignment in terms of facial shape and exact pose.

In contrast to the Faces data, Cars data has more diversity in viewpoints and car placement.
The fine-tuned models show different sizes, poses and backgrounds.
On the other hand, {\Name} produces consistently aligned cars.
We evaluate the viewpoint alignment with the regression model from Liao \etal~\cite{LiaoCVPR19} by measuring the mean difference in azimuth and elevation between Sedan and other domains. 
Fine-tuning achieves 53.2 and 3.8 degrees in azimuth and elevation, respectively. {\Name} achieves \textbf{21.0} and \textbf{2.2} degrees in azimuth and elevation, significantly outperforming the fine-tuning approach.
These results show that if domains have less diversity in poses and attributes, fine-tuning methods can produce reasonably aligned samples. 
However, as datasets become more diverse, it becomes challenging to enforce alignment without feature sharing.
{\Name} has the unique advantage of being a model that shares the \textit{same features} across domains to produce highly aligned samples while enabling a diverse set of applications.

\vspace{-5mm}
\section{Limitations}
\label{sec:limitations}
\vspace{-1mm}

We observed a slight deterioration in quality compared to StyleGAN2 on certain domains such as AFHQ-Cat (last column of Tab.~\ref{tab:low_data}).
As we have used the same core generator backbone~\cite{karras2020analyzing} for all experiments, future work includes improving the core generator, for example via increasing capacity to be more suitable for multi-domain modelling. 

For \textit{Cars}, {\Name} often puts vibrant colors on \textit{Sports Car} as it consists of mostly those colors. 
If one wishes to further impose stronger texture consistency among domains, one possible remedy is adding a regularization term encouraging each domain to output similar colors.

The morph maps in {\Name} are 2D and consequently cannot handle morphing in 3D.
If the geometric differences between domains have to be modelled in 3D, such as object rotations, {\Name} can only mimic them rather than performing true 3D morphing.
{\Name} is also not applicable for vastly different domains.
Lastly, each domain currently maintains a separate discriminator, which can be costly if the number of domains grows. We leave efficient discriminator training to future work.

\section{Conclusion}
\label{sec:conclusion}
\vspace{-1mm}

We introduced \textit{Polymorphic-GAN}, which produces aligned samples across multiple domains by learning the geometric differences through morph maps. 
{\Name}'s morph maps enable efficient sharing of generator features. This allows 
{\Name} to be utilized for diverse applications, including zero-shot segmentation transfer, image-to-image translation, image editing across multiple domains as well as training in low-data settings. PMGAN is the first GAN to efficiently synthesize aligned samples from multiple geometrically-varying domains at the same time. 

{\small
\bibliographystyle{ieee_fullname}
\bibliography{egbib}
}

\clearpage

\appendix
\addcontentsline{toc}{section}{Appendices}
\renewcommand{\thesection}{\Alph{section}}

\begingroup
\let\clearpage\relax
\onecolumn
\endgroup
\maketitle
\section*{\Large\selectfont{Supplementary Materials for Polymorphic-GAN: Generating Aligned Samples across Multiple Domains with Learned Morph Maps}}

\thispagestyle{empty}


\section{Model Architecture}
We provide additional descriptions of the architecture of {\Name} in this section.
\subsection{Pre-trained StyleGAN}
\label{sec:pretrained_stylegan_supp}
\vspace{-1mm}
{\Name} is composed of the pre-trained StyleGAN2's generator $G$, domain-specific morph layers $M^{1,...,N}$ and rendering layers $R^{1,...,N}$.
We first sample a noise vector $z \sim p(z)$ from the standard Normal prior distribution and feed it through $G$, which produces the output image $I^P$ and also the intermediate features $u_1, ..., u_L$ for $L$ features in $G$.
In this work, all experiments are carried out at $256\times 256$ RGB image resolution.
Thus, we store the generator features for each spatial resolution from $4\times4$ to $256\times256$ before the final features are transformed via a $1\times1$ convolution layer (\ie tRGB) that produces the output RGB values. The features are shaped as $(4\times4\times512)$, $(8\times8\times512)$, $(16\times16\times512)$, $(32\times32\times512)$, $(64\times64\times512)$, $(128\times128\times256)$ and $(256\times256\times128)$, where the first two dimensions correspond to the height and width, and the last dimension is for the number of channels.

\subsection{MorphNet}
\label{sec:morphnet_supp}
\vspace{-1mm}
Features $u_1, ..., u_L$ contain valuable information, including semantic content as well as fine-grained edge information.
We use these features to produce domain-specific morph maps that can modify the geometry embedded in the features to be suitable for each target domain.
The MorphNet component of {\Name} first reduces each feature map's channel dimension to be smaller through a $1\times1$ convolution layer and then upsamples all features to match the largest spatial resolution $256\times 256$. 
Each $1\times1$ convolution layers reduce the number of channels to 128 followed by a leaky ReLU~\cite{maas2013rectifier} activation function.

The upsampled features are concatenated channel-wise, resulting in a $(256\times256\times896)$ tensor.
It goes through two $3\times3$ convolution layers whose output channel dimensions are 512, followed by leaky ReLU. These layers are shared across domains and the spatial dimension is preserved (with stride=1 and padding=1). The conv layers and upsampling operations are represented as $\texttt{MergeFeatures}$ in Algorithm 1 in the main text.

We add a sinusoidal positional encoding~\cite{vaswani2017attention} for 2D to the merged features to inject grid position information which can be useful for learning geometric biases in a dataset.
We define the positional encoding as
\begin{align*}
&PE(x,y,4c) = sin(x / 10000^{8c/512}) \\
&PE(x,y,4c+1) = cos(x / 10000^{8c/512}) \\
&PE(x,y,4c+2) = sin(y / 10000^{(8c+4)/512}) \\
&PE(x,y,4c+3) = cos(y / 10000^{(8c+4)/512})
\end{align*}
where $x\in[0,255], y\in[0,255]$ for spatial dimensions, and $c\in[0,127]$ for the channel dimension.
  
Finally, this summed tensor is processed by domain-specific convolution layers $M^d$ for each domain $d$. 
$M^d$ is composed of two convolution layers. 
The first layer is spatial-dimension preserving  $3\times3$ conv layer that outputs 512 channels, followed by a leaky ReLU activation function.
The second layer is spatial-dimension preserving  $3\times3$ conv layer that outputs 2 channels, followed by a Tanh activation function and a scalar division by $\eta$ which is a hyperparameter that controls the maximum displacement we allow the morphing operation to produce. We use $\eta=3$ for all experiments in this paper.
Thus, $M^d$ produces a $H\times W\times2$ morph map $\mathcal{M}^d_\Delta$, normalized between $[-1/\eta, 1/\eta]$.
$\mathcal{M}^d_\Delta$ represents the relative horizontal and vertical direction that each pixel would get its value from (a pixel here is $(p,q)$ position in a 3-dim spatial tensor).

\subsection{Feature Morphing}
\label{sec:morphing_supp}
\vspace{-1mm}
We follow Spatial Transformer Networks (SPN)~\cite{jaderberg2015spatial} to differentiably morph features with $\mathcal{M}^d_\Delta$. 
We initialize a 2D sampling grid from an identity transformation matrix, normalized between $[-1, 1]$.
The sampling grid has the same shape as $\mathcal{M}^d_\Delta$, and each pixel $(p,q)$ in the sampling grid contains the absolute position $(x,y)$ of the source pixel that will be morphed into $(p,q)$. For example, if pixel $(p,q)$ has value $(-1, -1)$, the vector at the top left corner of the source feature map will be morphed into $(p,q)$.
The morph map $\mathcal{M}_{\Delta}^d$ is added to the grid, and we denote the resulting grid as $\Gamma \in \mathbb{R}^{H\times W\times 2}$. 
Unlike SPN that produces an affine transformation matrix with six parameters for sampling grid, we learn pixel-wise morphing maps, which gives us precise control for fine-detailed morphing. 
For each layer $l$ of generator features $\{u_1, ..., u_L\}_d$ from Section~\ref{sec:pretrained_stylegan_supp}, we perform the following \texttt{Morph} operation that bilinearly interpolates features: 
\begin{equation}
    \tilde{u}_l^{pq} = \sum_{n}^{H_l}\sum_{m}^{W_l} u_l^{nm}\text{max}(0, 1-|x^{pq} -m |)\text{max}(0,1-|y^{pq}-n|)
\end{equation}
where $\tilde{u}_l^{pq} \in \mathbb{R}^c$ is the morphed feature vector with $c$ channels at pixel $(p,q)$ for layer $l$, $u_l^{nm} \in \mathbb{R}^c$ is the source feature vector prior to \texttt{Morph} at pixel $(n,m)$ of $u_l \in \mathbb{R}^{H_l\times W_l\times c}$, and $(x^{pq}, y^{pq})$ is the sample point in $\Gamma$ for pixel $(p,q)$, assuming unnormalized grid coordinates for ease of presentation. 
Note that $\Gamma$ is also bilinearly interpolated to match the spatial dimension of each layer $(H_l, W_l)$.

The morphed features $\{\tilde{u}_1, ..., \tilde{u}_L\}_d$ are now geometrically transformed to be suitable for domain $d$.
Each of these features is then processed via further convolution layers $R^d$ to produce RGB images.
Each $R^d$ is composed of $L$ output heads for each morphed features in $\{\tilde{u}_1, ..., \tilde{u}_L\}_d$. 
Each head is implemented as three-layer modulated convolution layers from StyleGAN2~\cite{karras2020analyzing} which takes the feature $\tilde{u}_l$ as input. It also takes the latent code $w=\texttt{mapping}(z)$ as an additional input for the modulation process, where $\texttt{mapping}$ is the mapping layer of the core generator in $G$. The first two layers output 512 channels, followed by leaky ReLU activation, and the last layer outputs 3 RGB channels.
The RGB outputs from $L$ layers are summed together using skip connections as in StyleGAN2~\cite{karras2020analyzing}. 
Importantly, the $R^d$ layers can correct small unnatural distortions caused by the feature morphing process, in contrast to previous works that directly warp output images~\cite{shi2019warpgan}.

\section{Datasets}
\noindent We construct two multi-domain datasets for evaluation: 

\textbf{Cars} dataset consists of five classes of cars from the LSUN-Car dataset~\cite{yu15lsun}. We use an object classifier by Ridnik \etal~\cite{ridnik2021imagenet21k} that can output fine-grained object classes to divide the dataset into the following domains: Sedan (149K), SUV (52K), Sports car (58K), Van (25K), and Truck (22K), with the number of images in parentheses.
LSUN dataset is distributed from https://www.yf.io/p/lsun. 

\textbf{Faces} dataset consists of Flickr-Faces-HQ~\cite{karras2019style} (70K), MetFaces~\cite{Karras2020ada} (1.3K), as well as Cat (5.6K), Dog (5.2K) and Wild life (5.2K) from the AFHQ dataset~\cite{choi2020starganv2}.

FFHQ dataset is distributed from https://github.com/NVlabs/ffhq-dataset. The images are published in Flickr by their uploaders under either Creative Commons BY 2.0, Creative Commons BY-NC 2.0, Public Domain Mark 1.0, Public Domain CC0 1.0, or U.S. Government Works. The dataset itself is licensed under Creative Commons BY-NC-SA 4.0 license by NVIDIA Corporation.

MetFaces dataset is distributed from https://github.com/NVlabs/metfaces-dataset. The images are distributed under Creative Commons Zero (CC0) license by the Metropolitan Museum of Art. The dataset itself is licensed under Creative Commons BY-NC 2.0 license by NVIDIA Corporation.

AFHQ dataset is distributed from https://github.com/clovaai/stargan-v2. We use the original version of the dataset. The dataset is licensed under Creative Commons BY-NC 4.0 license by NAVER Corporation.

\clearpage
\begin{figure}
  \centering
   \includegraphics[width=1.0\linewidth]{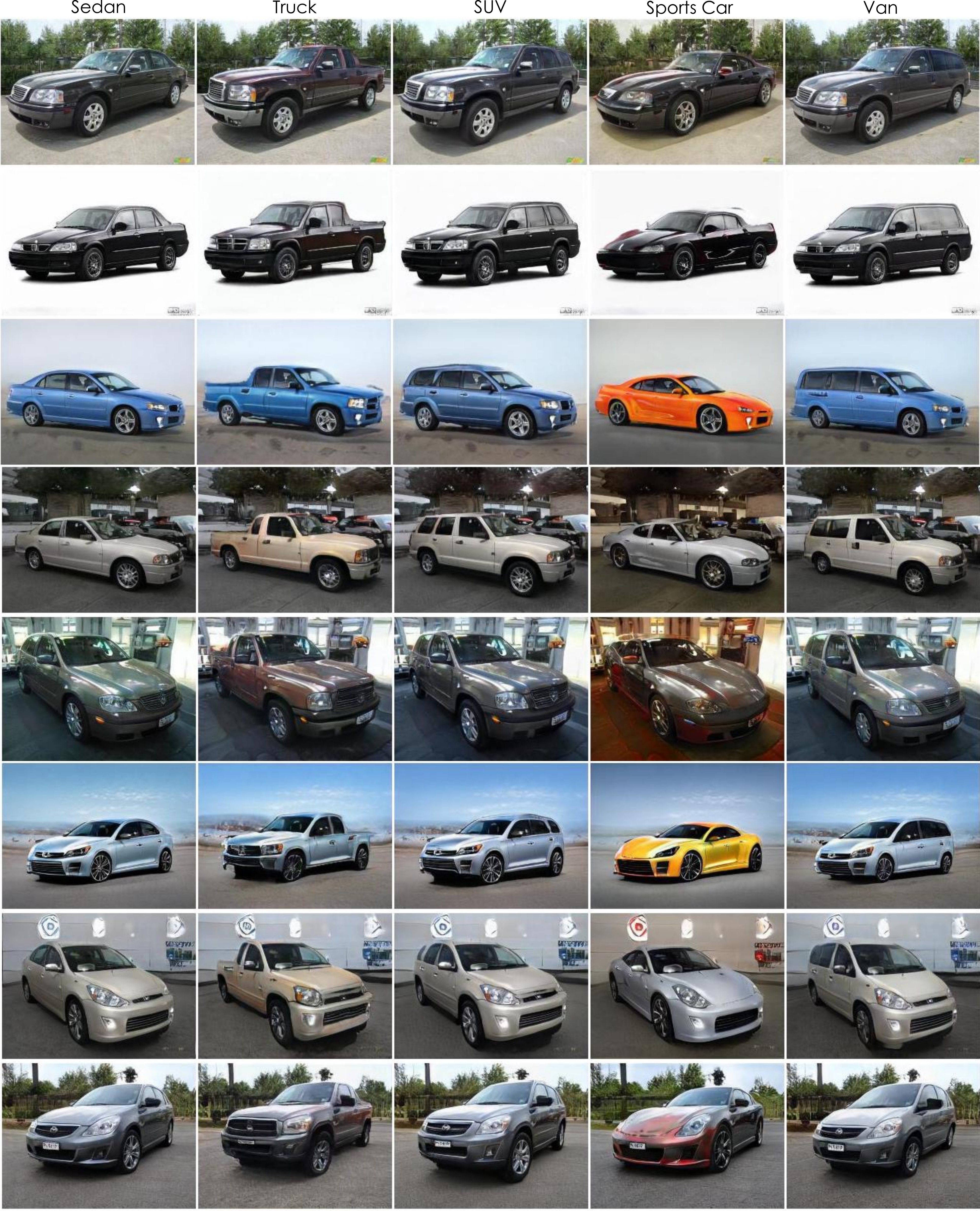}
   \caption{Aligned Samples from {\Name} trained on Cars dataset.}
   \label{fig:supp_lsun}
\end{figure}

\clearpage
\begin{figure}[H]
  \centering
   \includegraphics[width=0.85\linewidth]{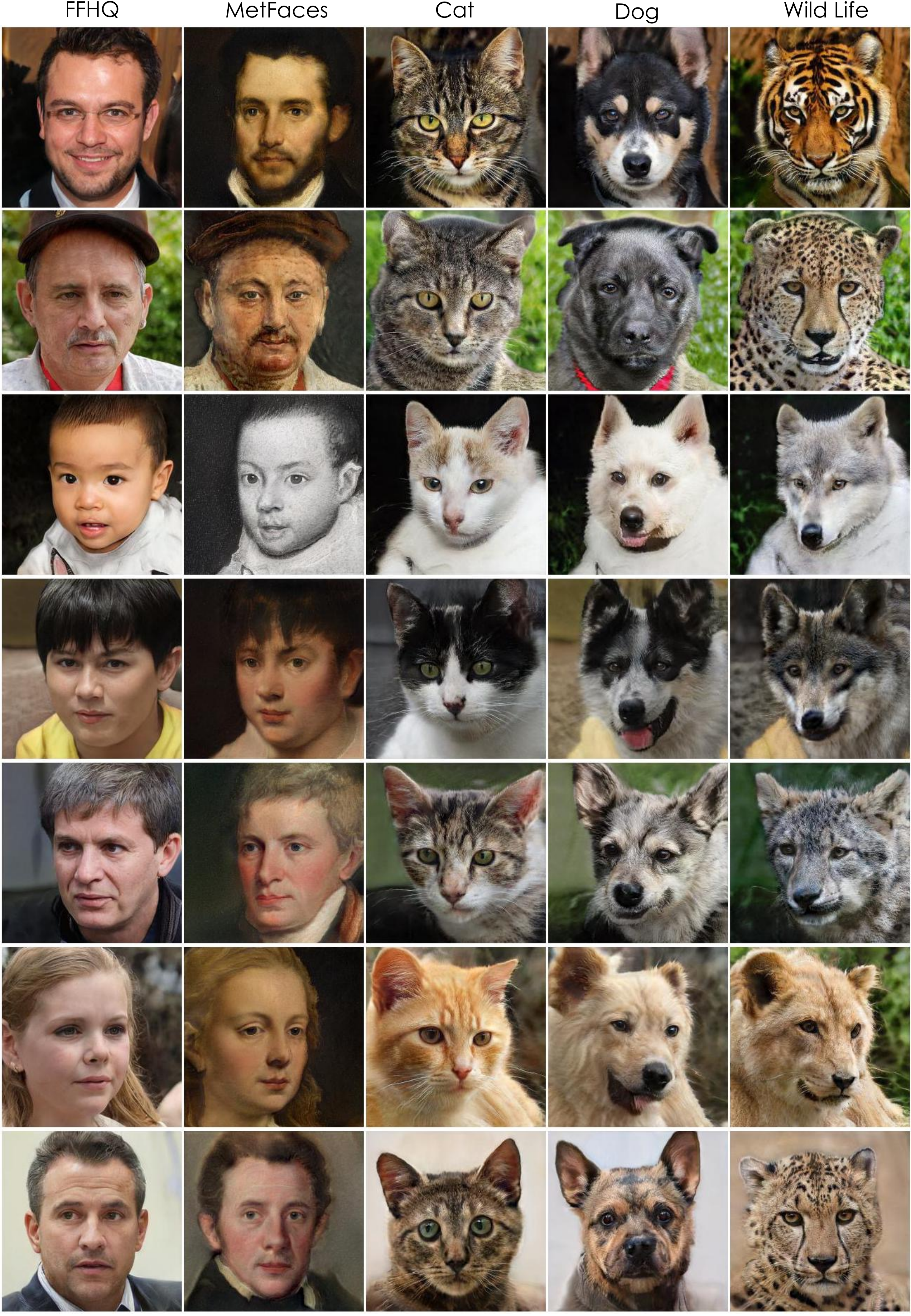}
   \caption{Aligned Samples from {\Name} trained on Faces dataset.}
   \label{fig:supp_faces}
\end{figure}

\clearpage
\section{Experiments}
In this section, we provide additional details on each of the models and algorithms used in the experiments section.

\begin{figure}[t]
\vspace{-1mm}
\begin{center}
    \includegraphics[width=0.79\linewidth]{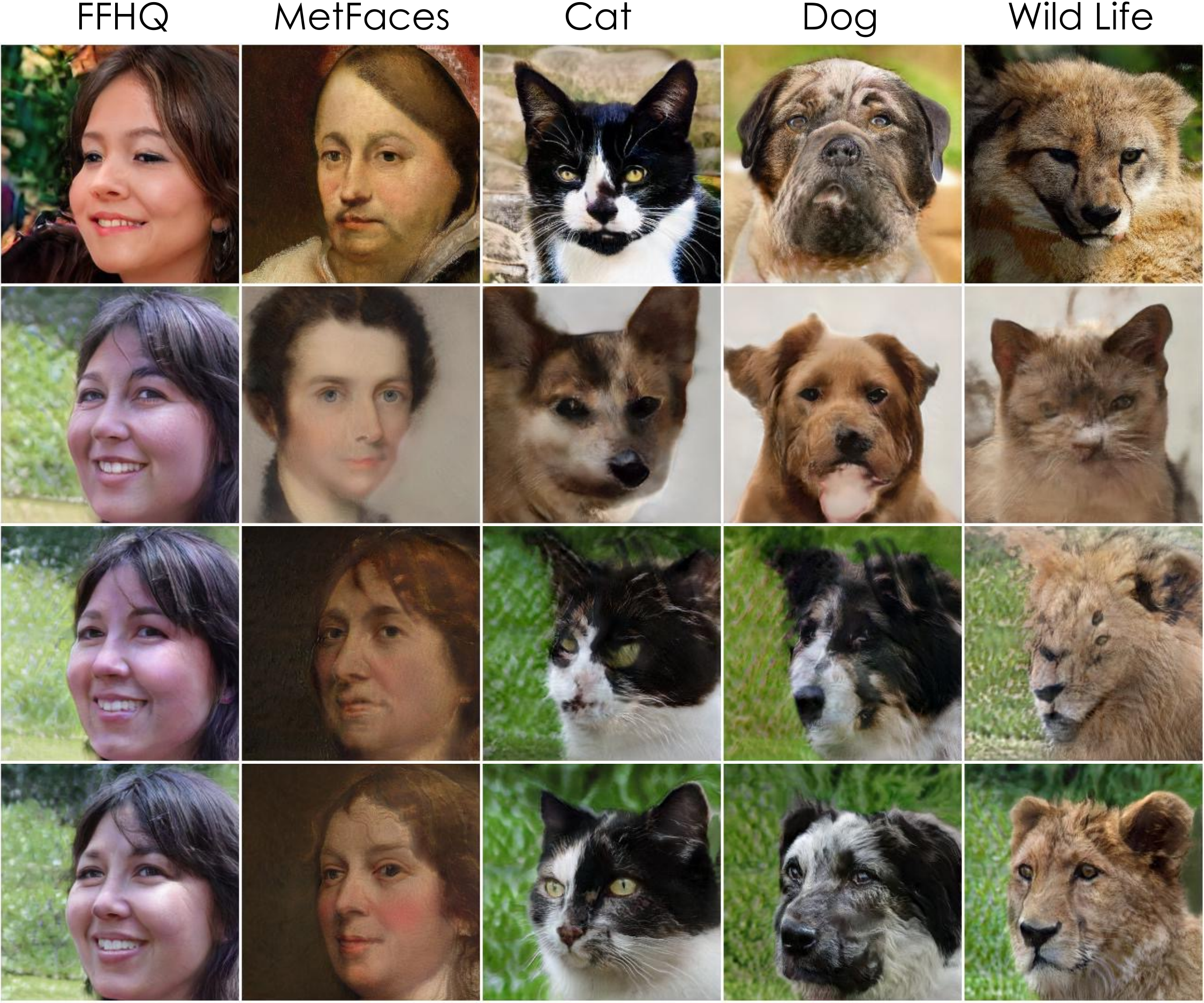}
\end{center}
\vspace{-6mm}
   \caption{\small  \emph{Top row}: DC-StyleGAN2, \emph{Second row}: *DC-StyleGAN2, \emph{Third row}: Ours without MorphNet, \emph{Last row}: Ours.}
   \vspace{-2mm}
\label{fig:sample2}
\end{figure}

\subsection{Ablation Studies}
\textit{Domain-Conditional StyleGAN2 (DC-StyleGAN2)} is a modified StyleGAN2 model that takes a one-hot encoded domain vector as an input. 
The 5-dimensional one-hot vector is embedded through a linear layer that outputs a 512-dimensional embedding vector. Then, the embedding is concatenated with $w$ latent from the mapping network $w=\texttt{mapping}(z)$, and then is fed through a linear layer that finally produces a 512-dimensional vector that goes through the generator.
The discriminator is the same as StyleGAN2's discriminator except that it is also conditioned on domain similar to the discriminator architecture of class-conditional BigGAN~\cite{brock2018large}. We add the dot product of the penultimate layer's output and domain embedding to the unconditioned output of the discriminator.
We provide additional aligned samples in Figure~\ref{fig:supp_lsun} and Figure~\ref{fig:supp_faces}. Figure~\ref{fig:sample2} provides an additional comparison with baselines. Except for DC-StyleGAN2 which does not share the same parent model, other models show samples from the same latent code.

We use domain classifiers to measure the domain classification accuracy indicating if models produce corresponding samples for each domain. 
They are implemented as ResNet-18~\cite{he2016deep} for the 5-way classification task, achieving 90.0\% and 99.9\% accuracy for Cars and Faces, respectively.
We note that classification is much easier for Faces because of their distinct texture.

\begin{figure}
  \centering
   \includegraphics[width=0.9\linewidth]{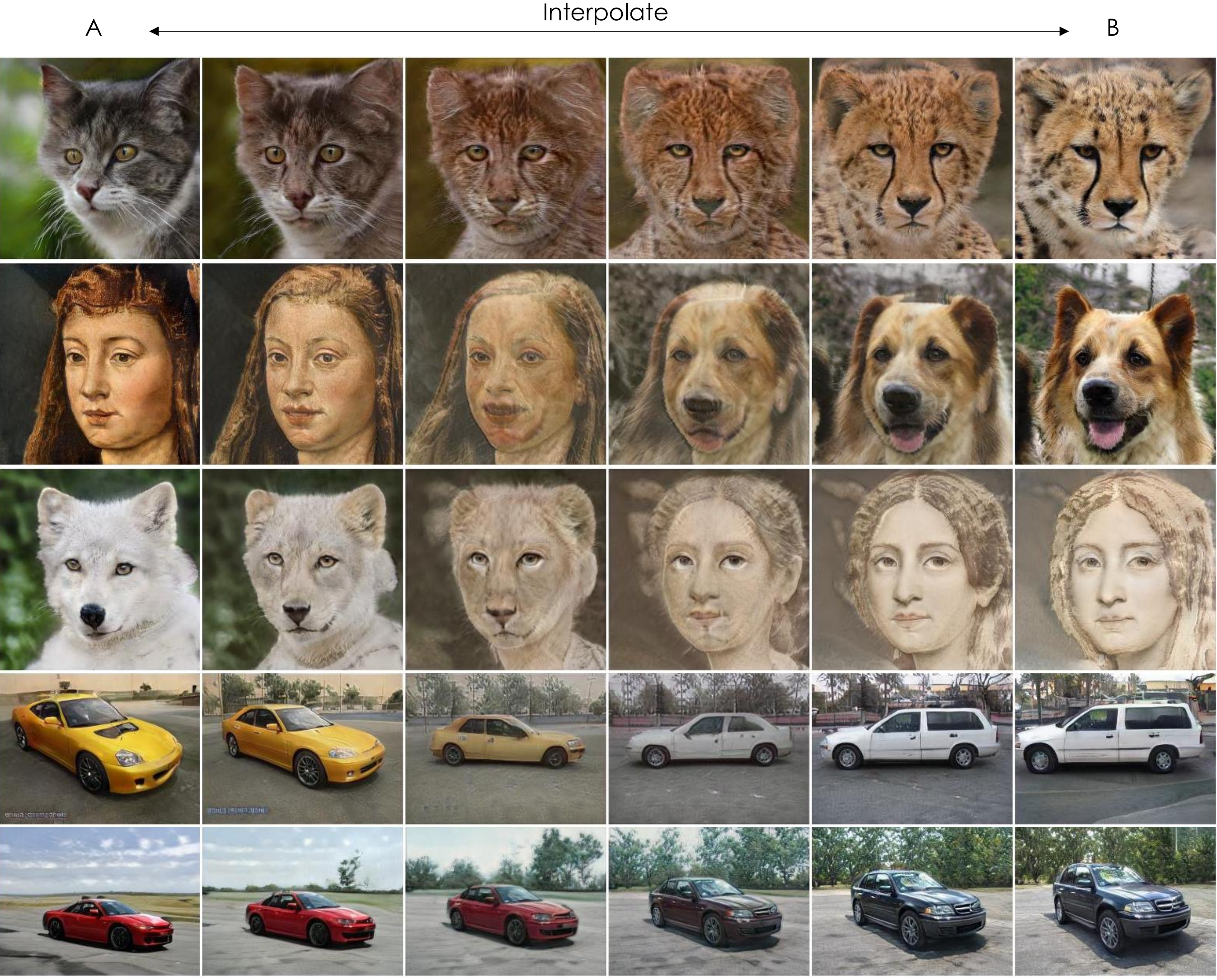}
   \caption{Cross-Domain Interpolation: we interpolate both the weights of domain-specific layers of two domains and their latent vectors A\&B.}
   \label{fig:supp_interp}
\end{figure}

\subsection{Morph Map and Edit Vector Transfer}
\begin{figure}
  \centering
   \includegraphics[width=0.88\linewidth]{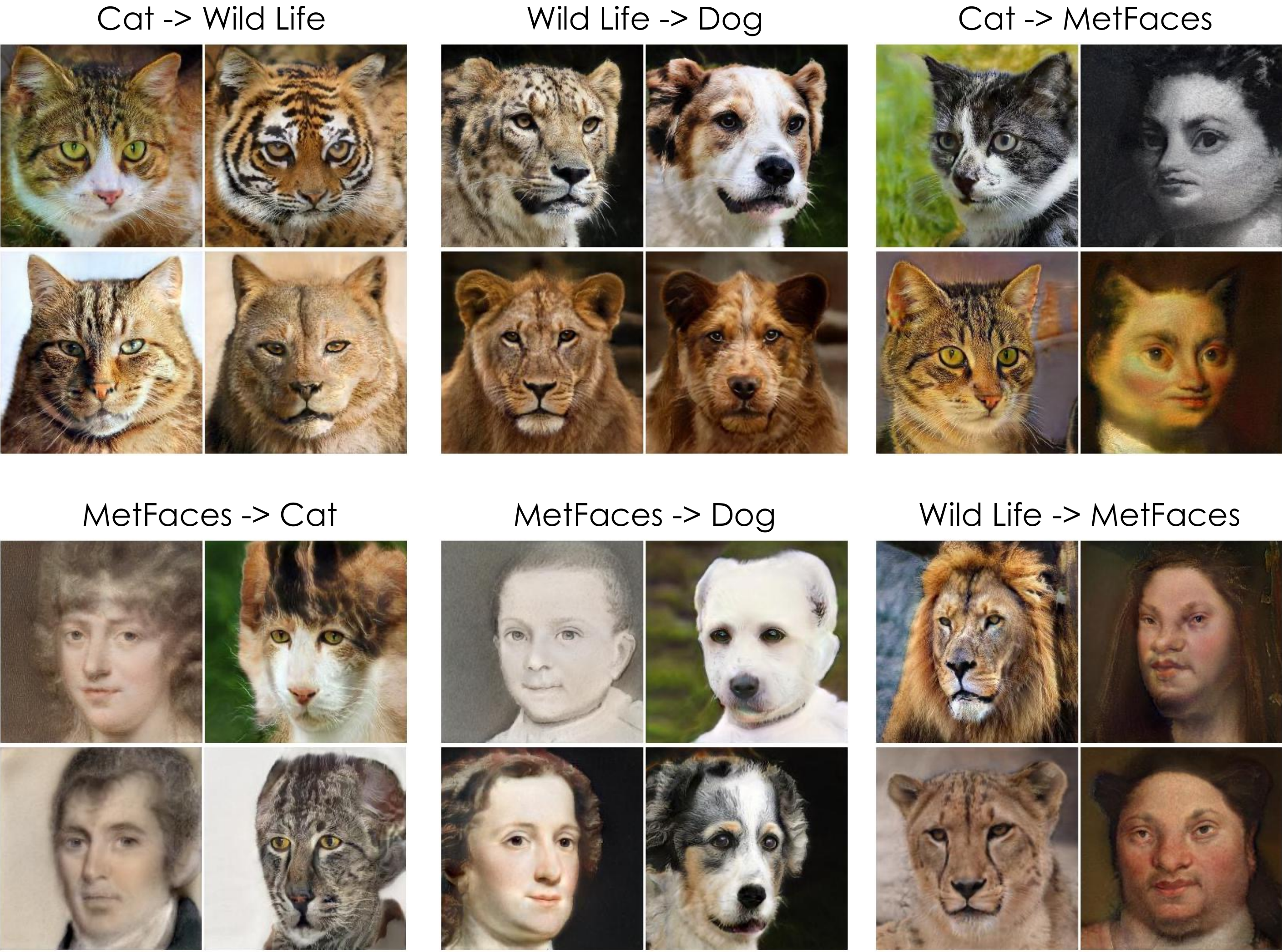}
   \caption{Rendering with the target domain's rendering layers while using the source domain's morph maps.}
   \label{fig:supp_fix_warp}
\end{figure}

\begin{figure}
  \centering
   \includegraphics[width=0.88\linewidth]{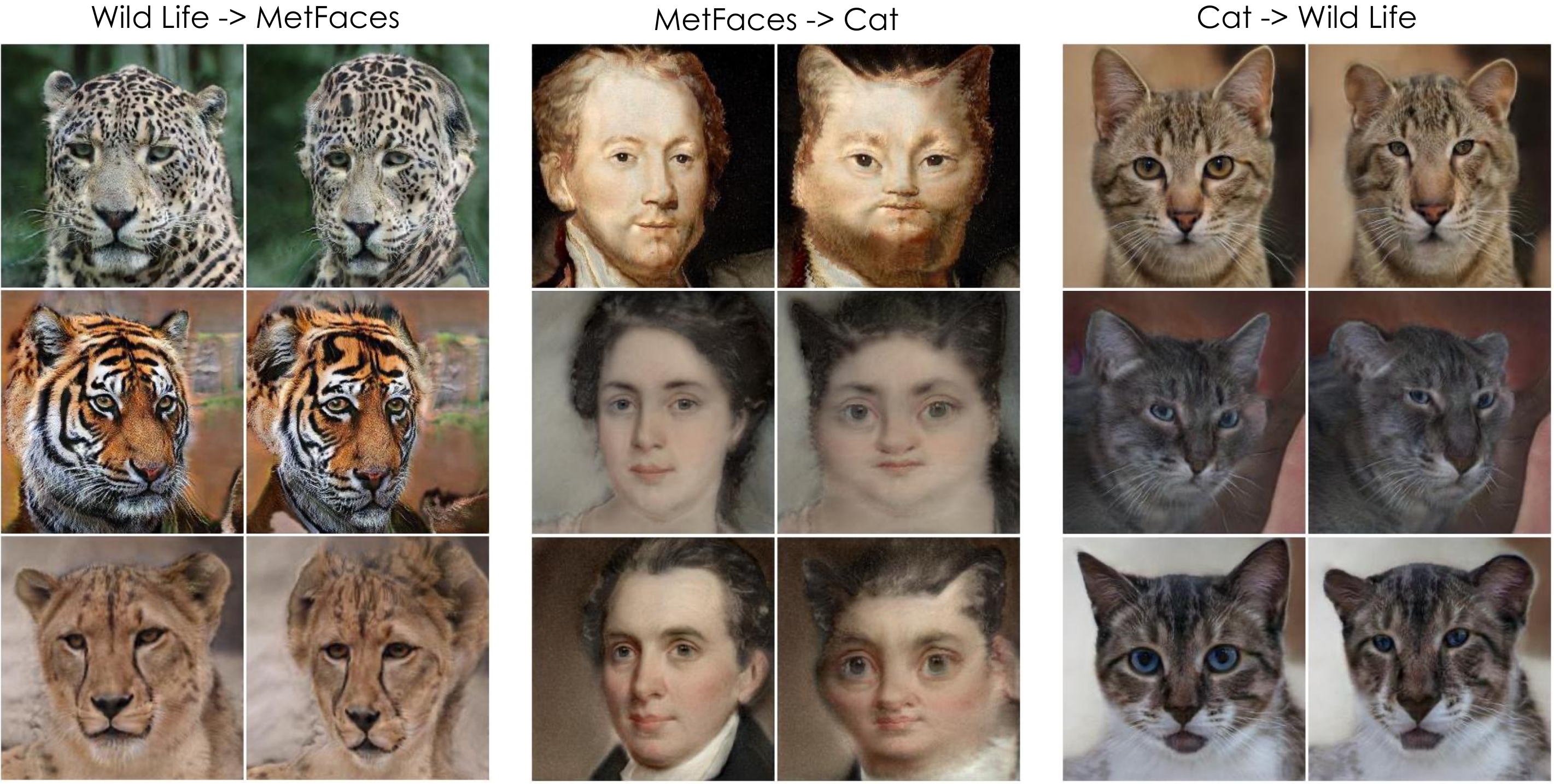}
   \caption{Rendering with the source domain's rendering layers while using the target domains morph maps.  Note how only the shape changes according to the target domain indicating the disentanglement between shape and rendering.}
   \label{fig:supp_fix_render}
\end{figure}

We provide additional examples on cross domain interpolation in Figure~\ref{fig:supp_interp}. 
Figure~\ref{fig:supp_fix_warp} shows translation between a source and target domain where we fix the morph map of the source domain and use target domain's rendering layers.
Figure~\ref{fig:supp_fix_render} also shows translations between two domains, but this time, the rendering layers of the source domain are kept fixed while the morph map from the target domain is used.
They show how {\Name} is able to produce novel outputs by disentangling the shape and texture with morph maps.

\begin{figure}
  \centering
   \includegraphics[width=1.0\linewidth]{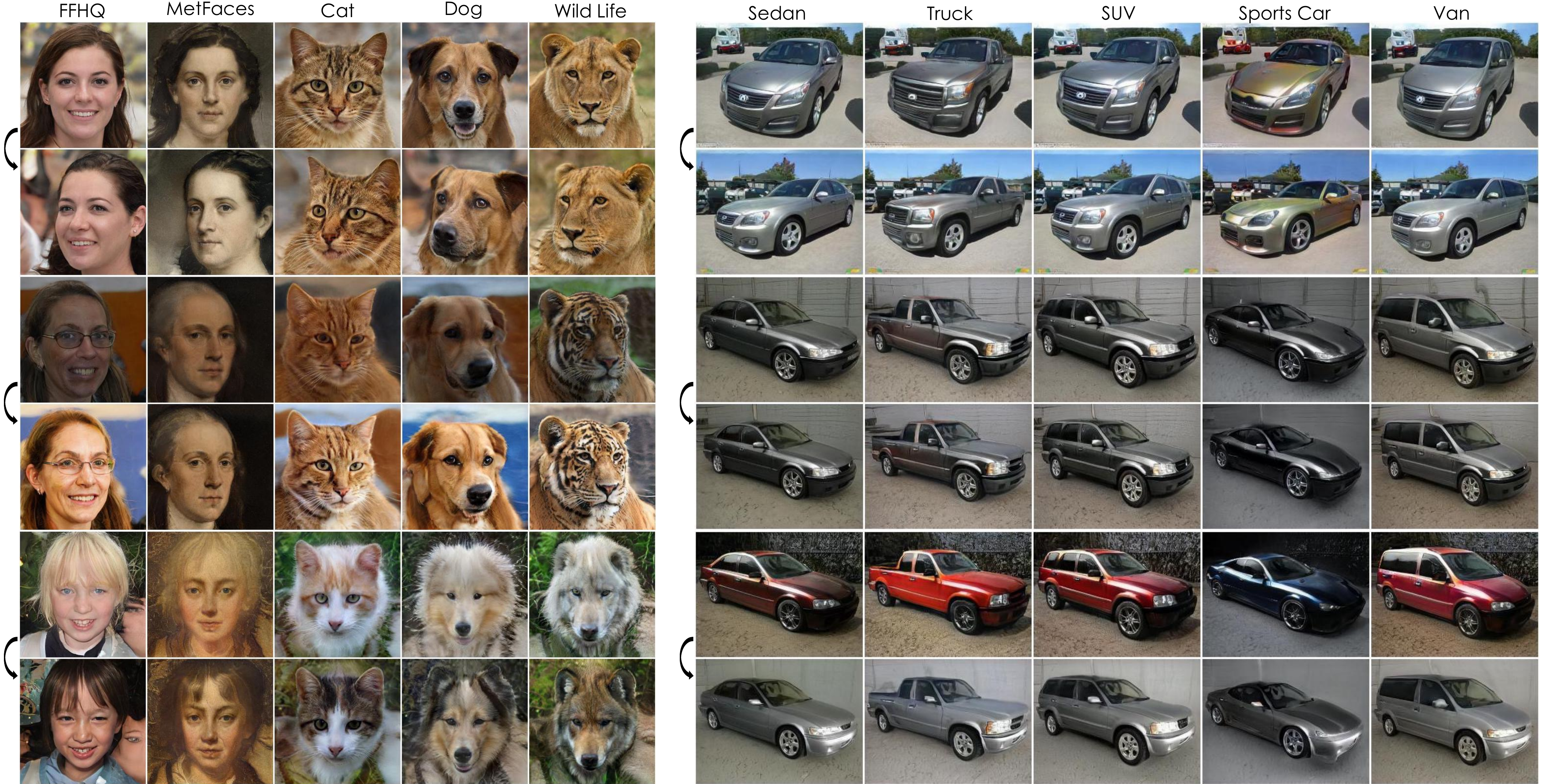}
   \caption{Edit transfer. Edit directions discovered through {\Name}'s core generator can be transferred across all domains. \textbf{Faces:} \textit{top}-rotation, \textit{middle}-brightness, \textit{bottom}-color. \textbf{Cars:} \textit{top}-rotation, \textit{middle}-zoom, \textit{bottom}-color.}
   \label{fig:edit_supp}
\end{figure}

For edit transfer, we use SeFa~\cite{shen2021closed} for its simplicity to find edit vectors in {\Name}.
SeFa produces edit directions in an unsupervised way by finding the eigenvectors of $A^TA$ where $A$ is the weight matrix of style layers in the core generator. Therefore, it is data independent and takes less than one second to find the edit vectors. We use the official implementation from https://github.com/genforce/sefa.
We find meaningful vectors such as rotation, zoom, lighting and elevation.
Figure~\ref{fig:edit_supp} contains additional examples of how edit vectors can be transferred across all domains for Faces and Cars datasets.

\subsection{Zero-shot Segmentation Transfer}
\begin{figure}
  \centering
   \includegraphics[width=1.0\linewidth]{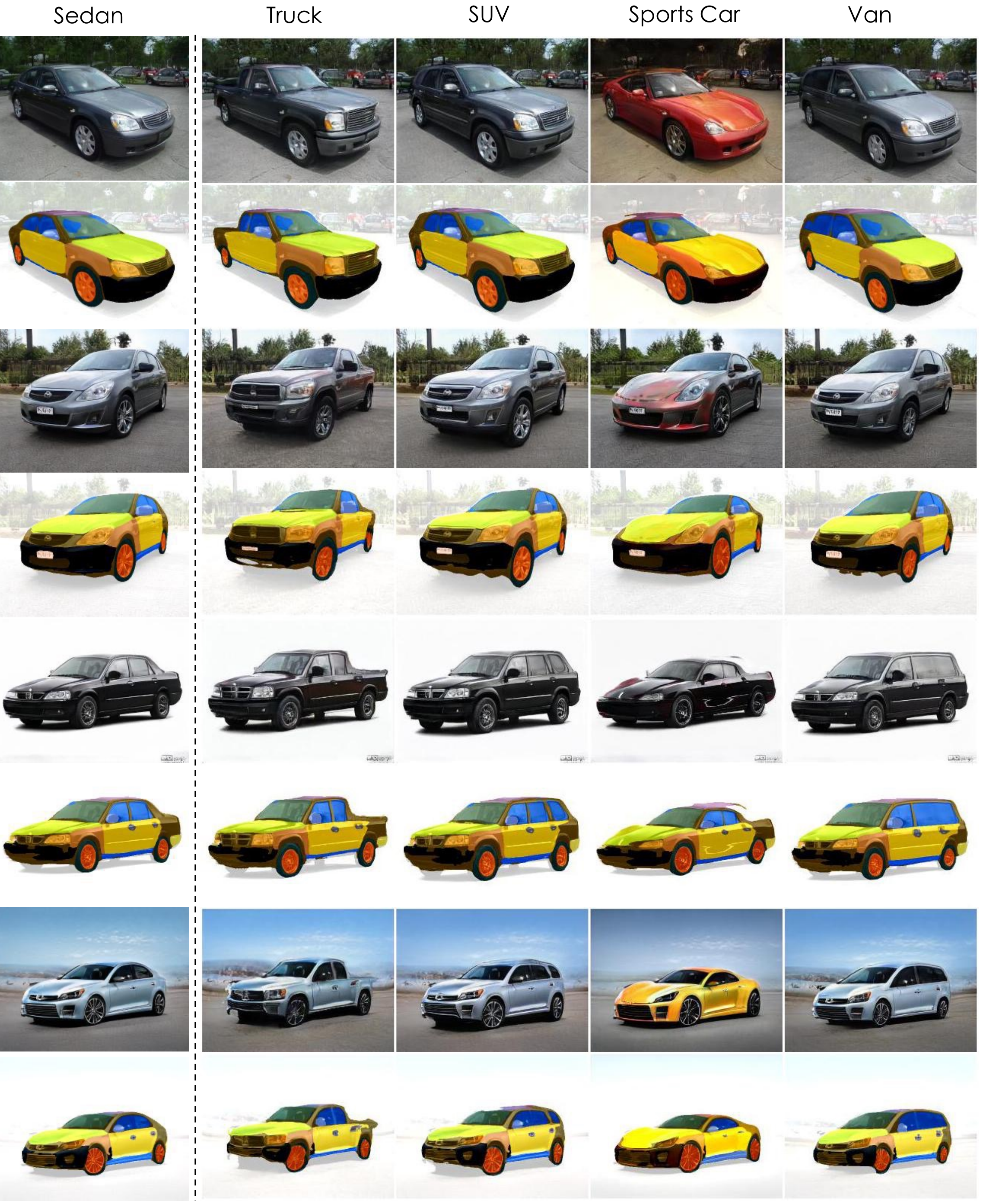}
   \caption{Zero-shot Segmentation Transfer on Cars. The segmentation mask from the leftmost column is transferred to all other domains.}
   \label{fig:supp_seg_lsun}
\end{figure}

\begin{figure}
  \centering
   \includegraphics[width=1.0\linewidth]{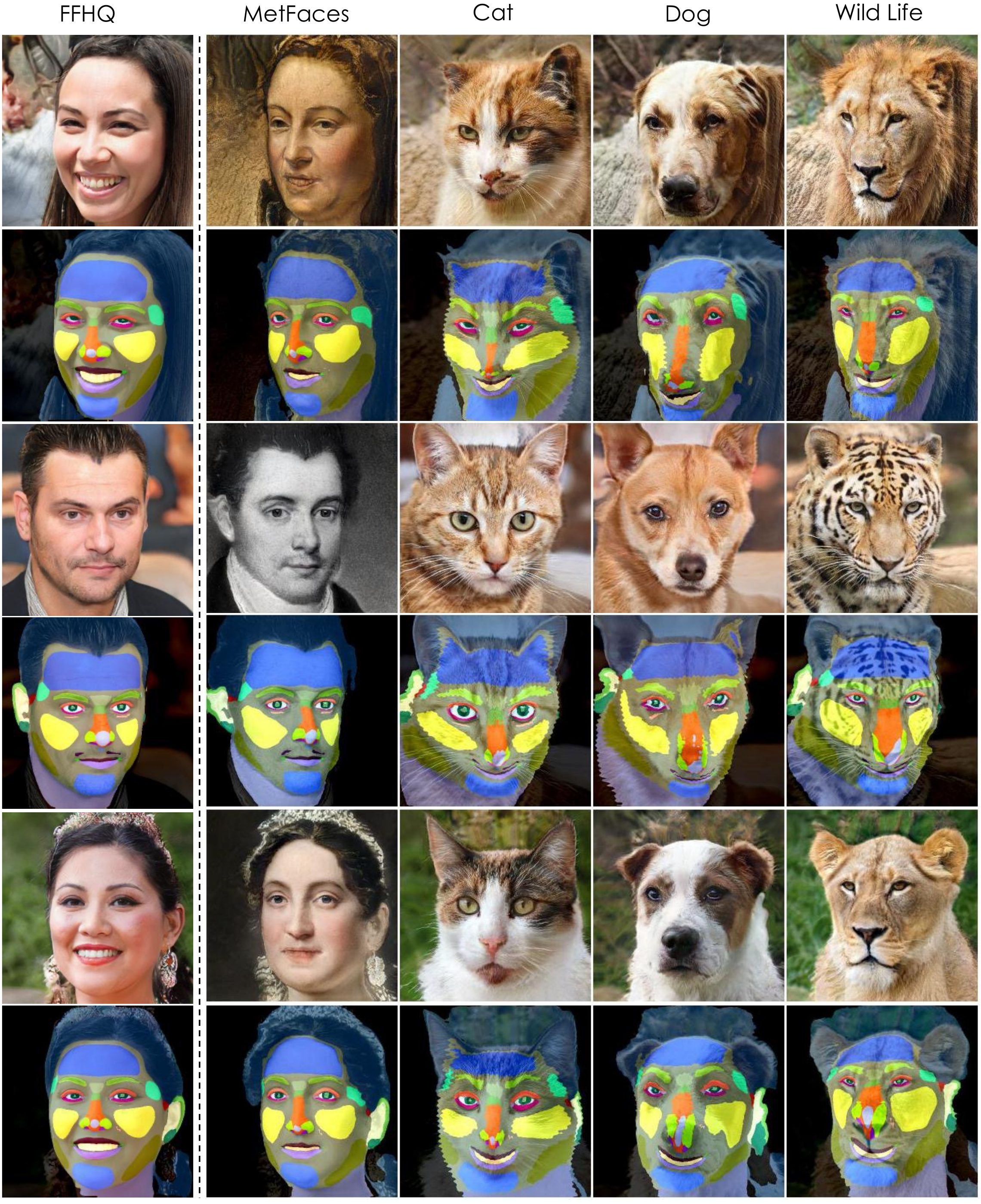}
   \caption{Zero-shot Segmentation Transfer on Faces. The segmentation mask from the leftmost column is transferred to all other domains.}
   \label{fig:supp_seg_faces}
\end{figure}

Assuming there exists a method that can output a segmentation map for images from the parent domain, it is possible to zero-shot transfer the segmentation mask to all other domains using {\Name}'s learned morph map.
We directly use the \texttt{Morph} operation on the segmentation map with $\mathcal{M}_\Delta$ after bilinearly interpolating the morph map to match the size of the mask.
As the morph map $\mathcal{M}_\Delta$ captures the geometric differences between domains, we can successfully use $\mathcal{M}_\Delta$ to transfer the parent's segmentation masks across domains.
We use pre-trained deeplab segmentation networks~\cite{chen2017deeplab} from DatasetGAN~\cite{zhang21} for Sedan and FFHQ domains for Cars and Faces datasets, respectively.
We then transfer them to other domains.
Specifically, we use the 20-part car model and 34-part face model trained with synthesized labels from DatasetGAN.
Figure~\ref{fig:supp_seg_lsun} and \ref{fig:supp_seg_faces} show additional segmentation transfer results. 
For Faces, we note that the noses of animals are always registered at the same location as the mouths of human domains. 
{\Name}'s 2D morph maps are interpretable and easy to edit as we know exactly what each pixel of the morph maps represent - the position of the source pixel that will be morphed into the pixel.
Therefore, to compensate for the fixed difference between the nose locations, we add a gaussian shaped downward offsets to the location of human face's nose before transferring its segmentation.
We set the peak of offset to be -0.15 in the vertical direction (0 for the horizontal direction) which corresponds to moving nose downward by 7.5\% of the image size. 
The offest only needs to be calculated once.
We emphasize that this still promotes feature sharing, as the rendering layers need to render the same features (\ie features representing \textit{mouth} from human domain) according to their domain.
This also shows editing shapes directly using morph map is an interesting direction, which we leave for future work.

\begin{figure}[t]
\begin{center}
    \includegraphics[width=1.0\linewidth]{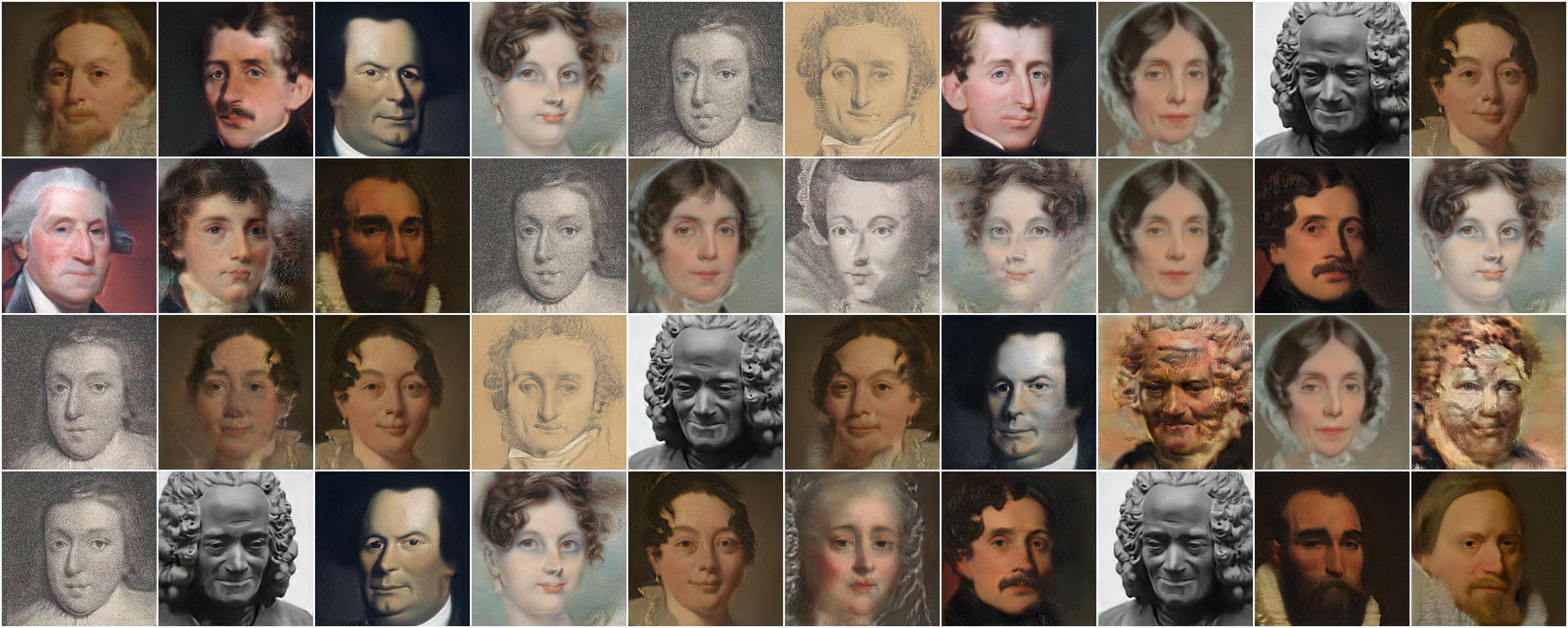}
\end{center}
   \caption{\small Random samples produced by StyleGAN2 trained on 5\% of MetFaces dataset from Table 7 in the main paper. }

\label{fig:mode_collapse}
\end{figure}

\subsection{Image-to-Image Translation}
Once an image is inverted in the latent space, {\Name} can naturally be used for image-to-image translation (I2I) tasks by synthesizing every other domain with the same latent code.
For both datasets, we use the w-latent space of StyleGAN which is the output space of the mapping network of the core generator.
On Faces dataset, we use encoder4editing~\cite{tov2021designing} (official code from https://github.com/omertov/encoder4editing) with an additional latent space loss we found to be helpful. The latent space loss is defined as $\sum_d \mathbb{E_d}_{w\sim p(w)} \lVert E_d(G_d(w)) - w \rVert$ where $E_d$ is the encoder for domain $d$ to be learned, $G_d$ is the fixed pre-trained {\Name} for domain $d$ and $w$ is the sampled latent vector. As we can synthesize as many sampled image and latent pair $(G(w), w)$ as we want, this loss helps stabilizing the encoder training. We add the latent space loss to the original loss function of encoder4editing.

On Cars dataset, we use latent optimization~\cite{karras2019style} to encode input images, which we found to be encoding better than encoder4editing. We suspect the diversity of Cars dataset makes learning a generic encoder for the dataset challenging.
We use 250 optimization steps with learning rate of 0.1, reducing the LPIPS~\cite{zhang2018perceptual} distance between the output and input images.

\begin{figure}
  \centering
   \includegraphics[width=0.95\linewidth]{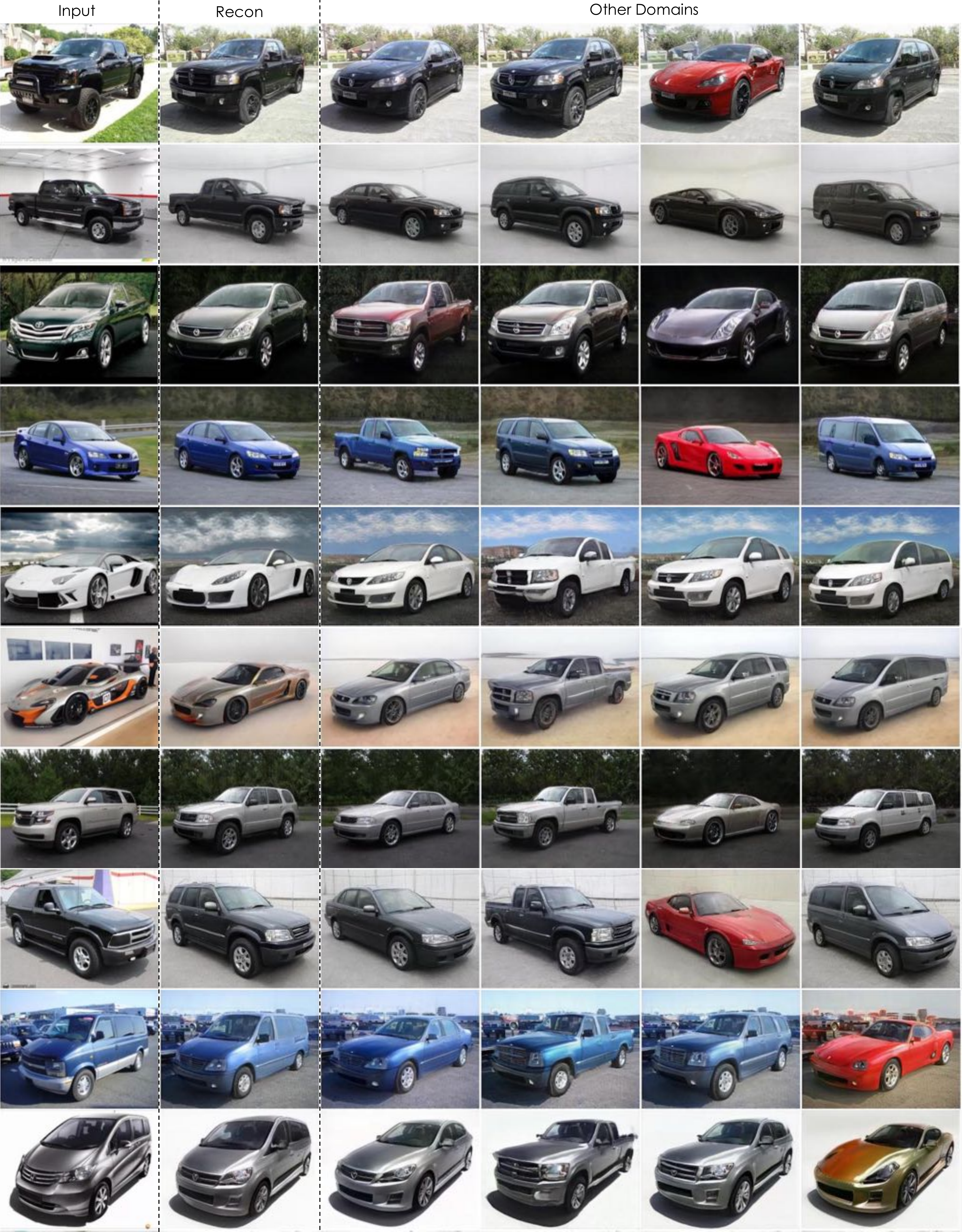}
   \caption{Image-to-Image translation results on Cars dataset.}
   \label{fig:supp_i2i_cars}
\end{figure}

\begin{figure}
  \centering
   \includegraphics[width=0.725\linewidth]{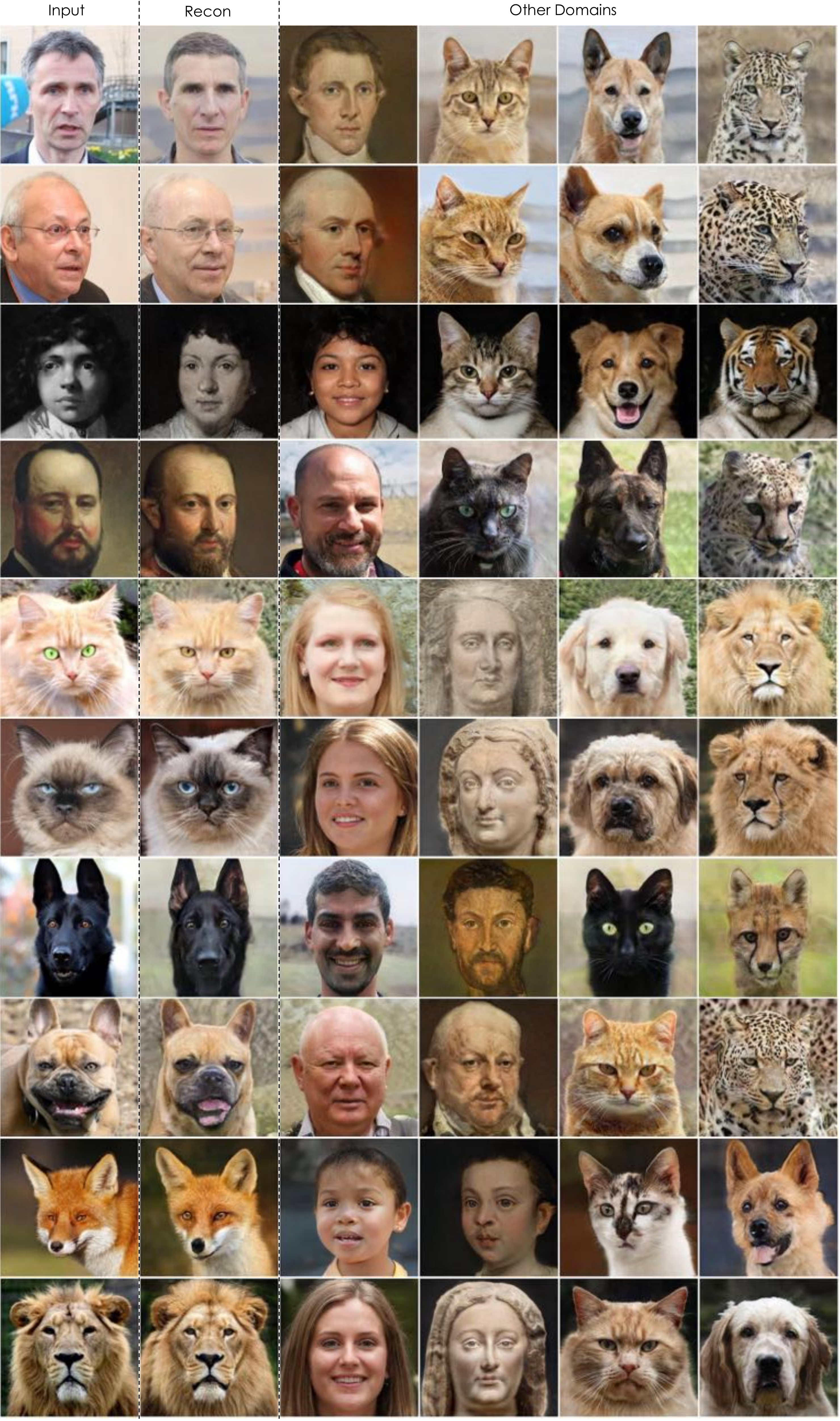}
   \caption{Image-to-Image translation results on Faces dataset.}
   \label{fig:supp_i2i_faces}
\end{figure}

\begin{figure}
  \centering
   \includegraphics[width=0.85\linewidth]{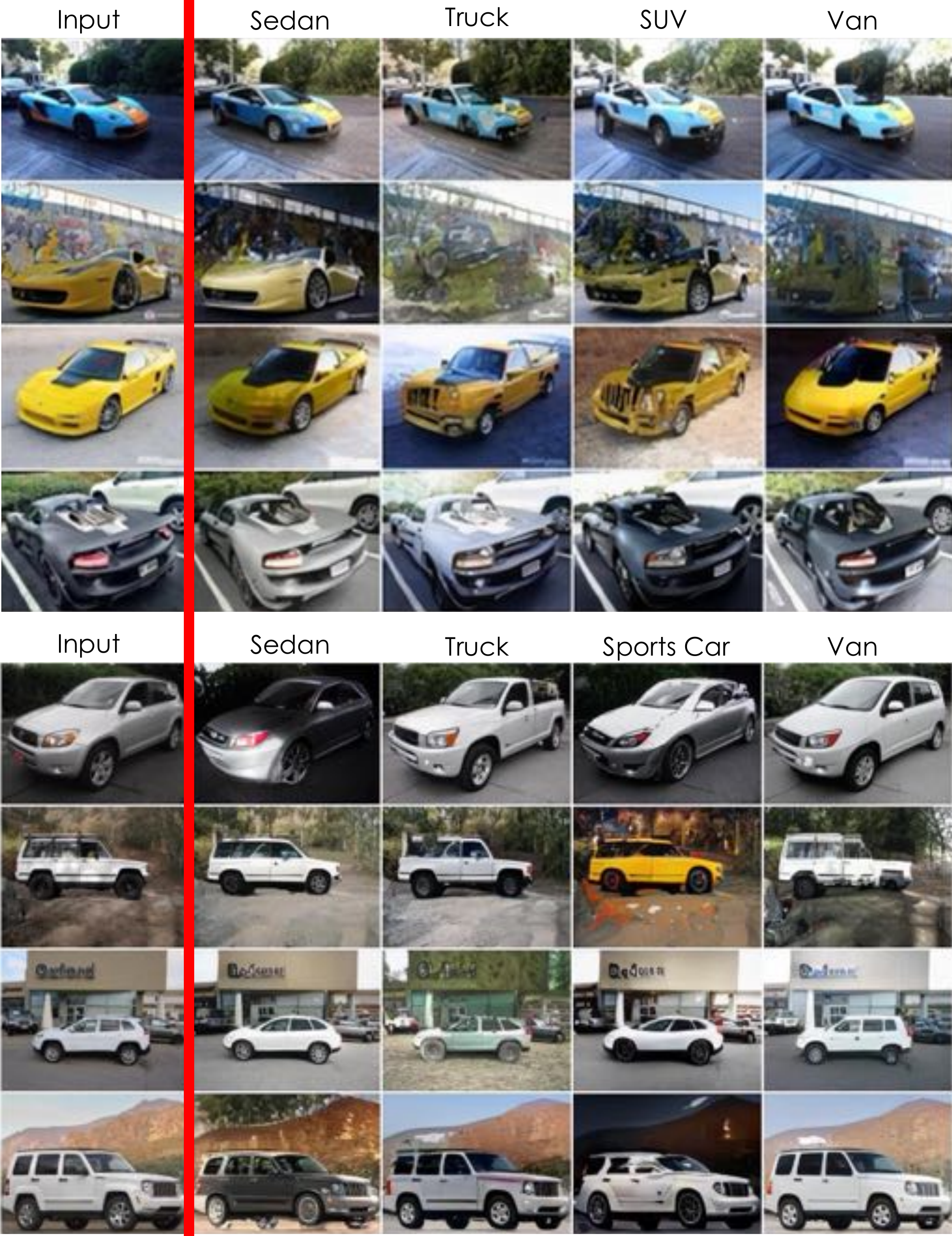}
   \caption{Additional Image-to-Image translation results on Cars dataset from baseline StarGANv2 model. StarGANv2 generally has difficulty changing the geometry of the input.}
   \label{fig:supp_i2_stargan}
\end{figure}

Figure~\ref{fig:supp_i2i_cars} and \ref{fig:supp_i2i_faces} contain additional image-to-image translation results from {\Name}. We also provide more translation results from StarGANv2 in Figure~\ref{fig:supp_i2_stargan}.

\subsection{Low-Data Regime}
As indicated in Section 3.5 of the main text, for low-data regime training, we weigh losses by $|\pi^d|/\text{max}_l|\pi^l|$ where $|\pi^d|$ is the number of training examples in domain $d$.
The intuition is that we want the generator features to be mostly learned from data-rich domains while 
domains with significantly less data leverage the rich representation with domain-specific layers.
To demonstrate how FID was not able to
capture the mode-collapse phenomenon, we include Figure~\ref{fig:mode_collapse}
showing random samples produced by StyleGAN2 trained
on 5
identity, indicating the model produces high-quality faces
by memorizing them, but this mode-collapse was not re-
flected well in FID as shown in Table 7. We will add more
examples for different models in the supplementary.

\subsection{Comparison to Plain Fine-Tuning}

 We use FreezeD~\cite{mo2020freeze} for fine-tuning experiments. 
 Mo \etal~\cite{mo2020freeze} found that freezing low-level discriminator layers improves fine-tuning performance. We freeze three discriminator layers for fine-tuning results. 
 We note that StyleGAN-ADA~\cite{Karras2020ada} also uses FreezeD along with their proposed adaptive discriminator augmentation. We have not used the adaptive discriminator augmentation in this paper, but adding it to FreezeD or our model potentially would improve results, especially when the number of data in target domain is small.

\clearpage

\end{document}